\documentclass[twoside,11pt]{article}

\usepackage{jmlr2e}

\let\proof\relax

\usepackage{amsfonts}
\usepackage{amsmath,amsthm}
\usepackage{caption}
\usepackage{subcaption}
\usepackage{hyperref}
\usepackage{fancyhdr} 
\usepackage{natbib}
\usepackage{tikz}
\usepackage{tikzit}
\tikzstyle{new style 0}=[fill={rgb,255: red,10; green,26; blue,255}, draw=black, shape=circle]
\tikzstyle{new style 1}=[fill=yellow, draw=black, shape=circle]

\tikzstyle{redline}=[-, draw=red]
\tikzstyle{blackline}=[-]
\tikzstyle{grayarrow}=[draw={rgb,255: red,141; green,141; blue,141}, ->]
\tikzstyle{blackarrow}=[draw=black, ->]

\usepackage{tikz-cd}
\usetikzlibrary{calc}

\definecolor{mygreen}{rgb}{0.0, 0.5, 0.0}

\tikzset{none/.style={thick}}
\tikzset{simple/.style={thick}}

\tikzset{
    ncbar angle/.initial=90,
    ncbar/.style={
        to path=(\tikztostart)
        -- ($(\tikztostart)!#1!\pgfkeysvalueof{/tikz/ncbar angle}:(\tikztotarget)$) 
        -- ($(\tikztotarget)!($(\tikztostart)!#1!\pgfkeysvalueof{/tikz/ncbar angle}:(\tikztotarget)$)!\pgfkeysvalueof{/tikz/ncbar angle}:(\tikztostart)$) 
           \tikztonodes
        -- (\tikztotarget) 
    },
    ncbar/.default=0.5cm,
}

\hypersetup{
    colorlinks=true,
    linkcolor=black,
    filecolor=magenta,      
    urlcolor=black,
    citecolor=black,
    breaklinks=true,
}

\newtheorem{Theorem}{Theorem}[section]

\newtheorem{Lemma}[Theorem]{Lemma}
\newtheorem{Definition}[Theorem]{Definition}
\newtheorem{Corollary}[Theorem]{Corollary}
\newtheorem{Remark}[Theorem]{Remark}

\newcommand{\R}{\mathbb{R}}
\newcommand{\act}{\textbf{a}}
\newcommand{\Cmplx}{\mathbb{C}}
\newcommand{\Gt}{\widetilde{G}}
\newcommand{\Q}{\widetilde{Q}}

\newcommand{\E}{\mathcal{E}}
\newcommand{\V}{\mathcal{V}}
\newcommand{\VV}{\widetilde{\mathcal{V}}}

\newcommand{\Wxf}{\texttt{W}_x^f}

\newcommand{\maxx}{\operatorname{\textbf{max}}\nolimits}
\newcommand{\minn}{\operatorname{\textbf{min}}\nolimits}

\firstpageno{1}

\begin{document}

\title{The Representation Theory of Neural Networks}

\author{\name Marco Antonio Armenta \email marco.antonio.armenta.armenta@usherbrooke.ca \\
       \addr Department of Mathematics \\ Department of Computer Science\\
       Universit\'e of Sherbrooke\\
       2500 Boulevard de l'Universit\'e, Sherbrooke, QC, Canada.
       \AND
       \name Pierre-Marc Jodoin \email pierre-marc.jodoin@usherbrooke.ca \\
       \addr Department of Computer Science\\
       Universit\'e of Sherbrooke\\
       2500 Boulevard de l'Universit\'e, Sherbrooke, QC, Canada.}

\editor{-}

\maketitle

\begin{abstract}%   <- trailing '%' for backward compatibility of .sty file
%In deep learning, an optimization algorithm is used to train neural networks. After the training, the neural network does not look at the optimization algorithm to make a prediction. This means that everything learned during the optimization is contained in the neural network structure. An implication from this observation is that to theoretically understand deep learning, one has to understand both the structure of the network and the optimization process. In this work, we take previous implicit assumptions on the architectures of neural networks and we make them explicit to examine them with ideas and results coming from the mathematical theory of quiver representations.

In this work, we show that neural networks can be represented via the mathematical theory of quiver representations.  More specifically, we prove that a neural network is a quiver representation with activation functions, a mathematical object that we represent using a {\em  network quiver}.  Also, we show that network quivers gently adapt to common neural network concepts such as fully-connected layers, convolution operations, residual connections, batch normalization, pooling operations {\color{black} and even randomly wired neural networks}. We show that this mathematical representation is by no means an approximation of what neural networks are as it exactly matches reality. This interpretation is algebraic and can be studied with algebraic methods.

%the  data  in  terms  of  thearchitecture of the neural network

% We observe that whenever a neural network computes a forward pass, it is always used as a quiver representation together with activation functions, independently of the task. We define the necessary mathematical objects to make this explicit, providing a new language in which to discuss neural networks. This interpretation is algebraic and can be studied with algebraic methods.

We also provide a quiver representation model to understand how a neural network creates representations from the data. We show that a neural network saves the data as quiver representations, and maps it to a geometrical space called the {\em moduli space}, which is given in terms of the underlying oriented graph of the network, {\color{black} i.e., its \textit{quiver}}. This results as a consequence of our defined objects and of understanding how the neural network computes a prediction in a combinatorial and algebraic way.  

Overall, representing neural networks through the quiver representation theory leads to {\color{black}9} consequences {\color{black}and 4 inquiries for future research} that we believe are of great interest to better understand what neural networks are and how they work. % We prove that the algebraic objects coming from deep learning are well suited to be understood with abstract methods, by interpreting the network function and the quiver representations of the data in terms of algebraic diagrams.
\end{abstract}


\section{Introduction}

Neural networks have achieved unprecedented performances in almost every area where machine learning is applicable~\citep{raghu2020survey, LeCun15,Goodfellow-et-al-2016}.  {Throughout \color{black} its} history, {\color{black} computer science has} had several turning points with ground-breaking consequences that unleashed the power of neural networks. To name a few, one might regard the chain rule backpropagation~\citep{Rumelhart86}, the invention of convolutional layers~\citep{LeCun89} and recurrent models~\citep{Rumelhart86}, the advent of low-cost specialized parallel hardware (mostly GPUs)~\citep{Krizhevsky12} and the exponential growth of available training data as some of the most important factors behind today's success of neural networks.

Ironically, despite our understanding of every atomic element of a neural network and our capability to successfully train it, it is still difficult with today's formalism to understand {\em what} makes neural networks so effective.  As neural nets increase in size, the combinatorics between its weights and activation functions makes it impossible (at least today) to formally answer questions such as : (i) why neural networks [almost] always converge towards a global minima regardless of their initialization, the data it is trained on and the associated loss function? (ii) what is the true capacity of a neural net? (iii) what are the true generalization capabilities of a neural net?

One may hypothesize that the limited understanding of these fundamental concepts derives from the more or less formal representation that we have of these machines.  Since the '80s, neural nets have been mostly represented in two ways: (i) a cascade of non-linear atomic operations (be it, a series of neurons {\color{black}with their activation functions}, layers, convolution blocks, etc.) often represented graphically (e.g., Fig.3 by~\citet{He16}) and (ii) a point in an N dimensional Euclidean space (where N is the number of weights in the network) lying on the slope of a loss landscape that an optimizer ought to climb down~\citep{Li18}.

In this work, we propose a fundamentally different way to represent neural networks.  Based on quiver representation theory, we provide a new mathematical footing to represent neural networks as well as the data they process.  We show that this mathematical representation is by no means an approximation of what neural networks are as it tightly matches reality.  %As soon as we compute with a neural network it becomes a quiver representation together with activation functions, and as a consequence that pair of mathematical objects determines the computations performed by the neural network.

In this paper, we do not focus on how neural networks learn, but rather on the intrinsic properties of their architectures and their forward pass of data. Therefore providing new insights on how to understand neural networks.  Our mathematical formulation accounts for the wide variety of architectures there are, and also usages and behaviours of today's neural networks.  For this, we study the combinatorial and algebraic nature of neural networks by using ideas coming from the mathematical theory of quiver representations~\citep{Assem06,Schiffler14}. {\color{black}
Although this paper focuses on feed-forward networks, a combinatorial argument on recurrent neural networks can be made to apply our results to them: the cycles in recurrent neural networks are only applied a finite amount of times, and once unraveled they combinatorially become networks that feed information in a single direction with shared weights \citep{Bengio13}.
}

This paper is based on two observations that expose the algebraic nature of neural networks and how it is related to quiver representations:
\begin{itemize}
    \item[1.] {\color{black}When computing a prediction,} neural networks are quiver representations together with activation functions.
    \item[2.] The forward pass of data through the network is encoded as quiver representations.
\end{itemize}
Everything else in this work is a mathematical consequence of these two observations.  Our main contributions can be summarized by the following six items:
\begin{enumerate}
    \item We provide the first explicit link between representations of quivers and neural networks.
    \item We show that quiver representations gently adapt to common neural network concepts such as fully-connected layers, convolution operations, residual connections, batch normalization, pooling operations, {\color{black}and any feed-forward architecture, since this is a universal description of neural networks}.
    \item We prove that algebraic isomorphisms of neural networks preserve the network function and obtain, as a corollary, that ReLU networks are positive scale invariant~\citep{Dinh17,Meng18,Neyshabur15}.
    \item We present the theoretical interpretation of data in terms of the architecture of the neural network and of quiver representations.
    \item We mathematically formalize a modified version of the manifold hypothesis \citep{Bengio13,Goodfellow-et-al-2016} in terms of the {\color{black}combinatorial} architecture of the network.
    \item We provide constructions and results supporting existing intuitions in deep learning while discarding others, and bring new concepts to the table.
\end{enumerate}

\section{Previous work}

In the theoretical description of the deep neural optimization paradigm given by~\citet{Choromaska15}, the authors underline that ``\textit{clearly the model (neural net) contains several dependencies as one input is associated with many paths in the network. That poses a major theoretical problem in analyzing these models as it is unclear how to account for these dependencies}."  Interestingly, this is exactly what quiver representations are about~\citep{Assem06,Barot15,Schiffler14}.

While as far as we know, quiver representation theory has never been used to study neural networks, some authors have nonetheless used a sub-set of it, sometimes unbeknownst to them.  It is the case of the so-called {\em positive scale invariance} of ReLU networks which~\citet{Dinh17} used to mathematically prove that most notions of loss flatness cannot be used directly to explain generalization. This property of ReLU networks has also been used by~\citet{Neyshabur15} to improve the optimization of ReLU networks. In their paper, they propose the {\em Path-SGD} (stochastic gradient descent), which is an approximate gradient descent method with respect to a path-wise regularizer.  Also, \citet{Meng18} defined a space where points are ReLU networks with the same network function, which they use to find better gradient descent paths. In this paper (cf. Theorem~\ref{thm:netfunc} and Corollary~\ref{cor:pos_sca_inv}), we prove that positive scale invariance of ReLU networks is a property derived from the representation theory of neural networks that we present in the following sections. {\color{black}We interpret these results as evidence of the algebraic nature of neural networks, as they exactly match the basic definitions of representation theory (i.e., quiver representations and morphisms of quiver representations).}

\citet{Wood96} used group representation theory to account for symmetries in the layers of a neural network. Our mathematical approach is different since quiver representations are representations of algebras~\citep{Assem06} and not of groups. Besides, \citet{Wood96} present architectures that match mathematical objects with nice properties while we define the objects that model the computations of the neural network. We prove that quiver representations are more suited to study networks due to their combinatorial and algebraic nature.

\citet{Healy04} mathematically represent neural networks by objects called \textit{categories}.  However, as mentioned by the authors, their representation is an approximation of what neural nets are as they do not account for each of their atomic elements.  In contrast, our quiver representation approach includes every computation involved in a neural network, be it a neural operation (i.e., dot product + activation function), layer operations (fully connected, convolutional, pooling) as well as batch normalization.  As such, our representation {\color{black} is a universal description of neural networks, i.e., the results and consequences of this paper apply to all neural networks.}

%The following can hardly be overstated: we include absolutely all computations of a neural network in our analysis, and not even one more.

Quiver representations have been used to find lower-dimensional sub-space structures of datasets~\citep{Chindris20} without, however, any relation to neural networks. Our interpretation of data is orthogonal to this one since we look at how neural networks interpret the data in terms of every single computation they perform.

{\color{black}Following the discussion by} S. Arora in his 2018 ICML tutorial~\citep{Arora18vid} {\color{black}on the characteristics of a theory for deep learning}, our goal is {\color{black}precisely this. Namely,} to provide a theoretical footing that can validate {\color{black} and formalize} certain intuitions about deep neural nets and lead to new insights and new concepts.  One such intuition is related to feature map visualization.  It is well known that feature maps can be visualized into images showing the input signal characteristics and thus providing intuitions on the behavior of the network and its impact on an image~\citep{Yosinski15,Feghahati19}. This notion is strongly supported by our findings.  Namely, our data representation introduced in Section~\ref{sec:datarep} is a thin quiver representation that contains the network features (i.e., neuron outputs or feature maps) induced by the data.  Said otherwise, our data representation includes both the network structure and the {\color{black}neuron's inputs and outputs induced by a forward pass of a single data sample} (see Eq.~(\ref{Wxf}) in page \pageref{Wxf} and the proof of Theorem~\ref{thm:data}).  {\color{black}Our data quiver representations contain every feature map during a forward pass of data and so it is aligned with the notion of \textit{representations} in representation learning \citep{Bengio13, Goodfellow-et-al-2016, Hinton07}}.

We show in Section~\ref{sec:moduli} that our data representations lie into a so-called {\em moduli space}.  Interestingly, the dimension of the moduli space is the same value {\color{black}that was} computed by~\citet{Zheng19} and used to measure the capacity of ReLU networks. They empirically confirmed that the dimension of the moduli space is directly linked to generalization. {\color{black}Our results suggest that the findings mentioned above can be generalized to any neural network via representation theory.}

The moduli space also formalize a modified version of the manifold hypothesis for the data~\citep[see][chap.~5.11.3]{Goodfellow-et-al-2016}.  This hypothesis states that high-dimensional data (typically images and text) live on a thin and yet convoluted manifold in their original space. {\color{black}We show that this data manifold can be mapped to the moduli space while carrying the feature maps induced by the data, and then it is related to notions appearing in manifold learning, \citep{Bengio13, Hinton07}. {\color{black} Our results, therefore, create} a new bridge between the mathematical study of these moduli spaces \citep{Reineke08,Das19,Reineke20} and the study of the training dynamics of neural networks inside these moduli spaces}.

%While this assumption is usually considered true, no theorem {\color{black}has ever proven it}. Concerning our data quiver representations, which are derived from the neural net architecture and not by the data itself, we mathematically prove that they lie in a manifold (generated by the image of an explicitly provided map) inside the moduli space.    %Our results, however, work for any kind of activation functions and any kind of architectures.

%Inspired by the goal of a theory for deep learning of~\citet{Arora18vid}, we will make it explicit how our findings support certain intuitions people have had in regards of neural networks.

%There is also the notion of feature spaces~\citep{Yosinski15,Feghahati19} where one interprets the neuron values after a forward pass in a layer or a channel as coordinates of a space, that can also be visualized as images resembling some of the input image characteristics in the case of a well trained net.  This notion is strongly supported by our findings.  Namely, the coordinates of feature spaces get encoded into the quiver representations induced by the data, all at the same time, see Eq.~(\ref{Wxf}) in page \pageref{Wxf} and the proof of Theorem~\ref{thm:data}. 

Naive pruning of neural networks~\citep{Frankle18} where the smallest weights get pruned is also explained by our interpretation of the data and the moduli space (see consequence 4 on Section~\ref{cons:7:4}), since the coordinates of the data quiver representations inside the moduli space are given as a function of the weights of the network and the activation outputs of each neuron on a forward pass (cf. Eq.~(\ref{Wxf}) in page \pageref{Wxf}).

%It is empirically known that residual connections provide a solution to the vanishing and exploding gradient problem~\citep{He16}.  We prove that residual connections also give a better actualization of the coordinates of the data representations inside the moduli space (c.f. Consequence 3 in section \ref{sec:muduli_consequences}).  Also, we show that residual connections are positive scale invariant when considered as quiver representations as opposed to what what Meng et al previously hypothesized~\citet{Meng18}.% that there is no positive scale invariance across residual connections.   %(c.f Remark \ref{ex:res}).

%Our analysis of the computations of a neural network explains why the change of activation functions can drastically affect the expressiveness of a neural network~\citep{tanh,ReLU} (cf. construction \ref{Wxf} in page \pageref{Wxf}, and the discussion after Theorem~\ref{thm:data}).

There exist empirical results where, up to certain restrictions, the activation functions can be learned~\citep{Goyal19} and our interpretation of the data supports why this is a good idea in terms of the moduli space. For further details see Section~\ref{sec:7:cons}.

%It is known that max-pooling layers give small amount of translation invariance at each level since the precise location of the most active feature detector is thrown away, and this produces doubts about the use of max-pooling layers~\citep[cf.][]{Hinton14vid,Sabour17}.  An alternative to this is the use of attention-based pooling~\citep{Kosiorek19}, which is a global-average pooling.  Our interpretation provides a framework that supports why these doubts about the use of max-pooling layers exists: they break the algebraic structure of a neural network.  However, average pooling layers, and therefore global-average pooling layers, are perfectly consistent with respect to our results since they are given by fixed weights for any input vector (see Remark \ref{rmk:pooling}).

%There is also a current intuition that neural networks have very few levels of structure~\citep{Hinton14vid}, unlike other engineered systems. %And no wonder since we can only see neurons, layers and whole networks. Nevertheless, the main consequence of this work is that neural networks have a very rich algebraic structure.

\section{Preliminaries of Quiver Representations}

Before we show how neural networks are related to quiver representations, we start by defining the basic concepts of quiver representation theory~\citep{Assem06, Barot15, Schiffler14}. {\color{black}The reader can find a glossary with all the definitions introduced in this and the next chapters at the end of this paper.}

\begin{Definition} \citep[chap.~2]{Assem06}
    A \textbf{quiver} $Q$ is given by a tuple $(\V,\mathcal{E},s,t)$ where  $(\mathcal{V},\mathcal{E})$ is an oriented graph with a set of vertices $\mathcal{V}$ and a set of oriented edges $\mathcal{E}$, and maps $s,t:\mathcal{E} \to \mathcal{V}$ that send $\epsilon \in \mathcal{E}$ to its source vertex $s(\epsilon) \in \mathcal{V}$ and target vertex $t(\epsilon)\in \mathcal{V}$, respectively.
\end{Definition}

{\color{black}
Throughout the present paper, we work only with quivers whose sets of edges and vertices are finite. 

\begin{Definition} \cite[chap.~2]{Assem06}
      A \textbf{source vertex} of a quiver $Q$ is a vertex $v\in \V$ such that there are no oriented edges $\epsilon \in \mathcal{E}$ with target $t(\epsilon)=v$. A \textbf{sink vertex} of a quiver $Q$ is a vertex $v \in \V$ such that there are no oriented edges $\epsilon \in \mathcal{E}$ with source $s(\epsilon)=v$. A \textbf{loop} in a quiver $Q$ is an oriented edge $\epsilon$ such that $s(\epsilon)=t(\epsilon)$.
\end{Definition}}

\begin{figure} 
\centering
%\begin{subfigure}{.5\textwidth}
%  \centering
(a)  \includegraphics[width=.25\linewidth]{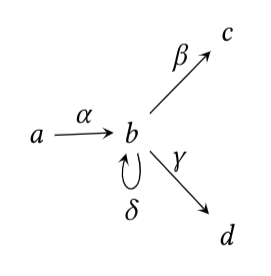}
%  \caption{A quiver.}
%\end{subfigure}%
%\begin{subfigure}{.5\textwidth}
%  \centering
(b)  \includegraphics[width=.25\linewidth]{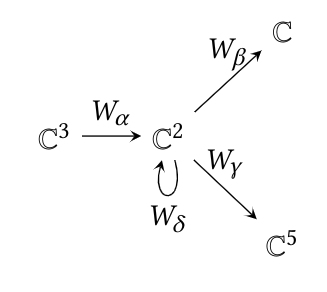}
%  \caption{A representation.}
%\end{subfigure}
%\begin{subfigure}{.5\textwidth}
%  \centering
(c)  \includegraphics[width=.25\linewidth]{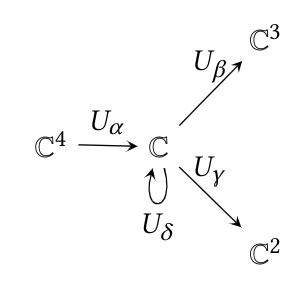}
%  \caption{Another representation.}
%\end{subfigure}
\caption{(a) A quiver $Q$ with vertices $\mathcal{V} = \{ a,b,c,d \}$ and oriented edges $\mathcal{E} = \{ \alpha, \beta, \gamma, \delta \}$, where the source and target maps are defined by $s(\alpha)=a$, $s(\beta)=b$, $s(\gamma)=b$, $s(\delta)=b$, $t(\alpha)=b$, $t(\beta)=c$, $t(\gamma)=d$ and $t(\delta)=b$. (b) A quiver representation $W$ over $Q$, where vertices $a$ to $d$ are complex 3D, 2D, 1D and 5D vector spaces, and $W_\alpha$ is a $2\times3$ matrix, $W_\beta$ is a $1\times2$ matrix, $W_\gamma$ is a $5\times2$ matrix and $W_\delta$ is a $2\times2$ matrix. (c) Another quiver representation $U$ over $Q$, where $a$ to $d$ are complex 4D, 1D, 3D and 2D vector spaces, and $U_\alpha$ is a $1\times4$ matrix, $U_\beta$ is a $3\times1$ matrix, $U_\gamma$ is a $2\times1$ matrix and $U_\delta$ is a $1\times1$ matrix.} \label{fig:Quiver_Reps}
\end{figure}

\begin{Definition}  \citep[chap.~3]{Assem06}
    If $Q$ is a quiver, a \textbf{quiver representation} of $Q$ is given by a pair of sets 
    \[
        W := \big( (W_v)_{v\in \V}, (W_\epsilon)_{\epsilon \in \E} \big)
    \]
    where the $W_v$'s are vector spaces indexed by the vertices of $Q$, and the $W_\epsilon$'s are linear maps indexed by the oriented edges of $Q$, such that for every edge $\epsilon \in \E$
    \[
        W_\epsilon : W_{s(\epsilon)} \to W_{t(\epsilon)}.
    \]
\end{Definition}
Fig.~\ref{fig:Quiver_Reps}(a) illustrates a quiver $Q$ while Fig.~\ref{fig:Quiver_Reps}(b),(c) are two quiver representations of $Q$.

\begin{figure} 
%\centering
%\begin{subfigure}{.4\textwidth}
%  \centering
(a)  \includegraphics[width=.3\linewidth]{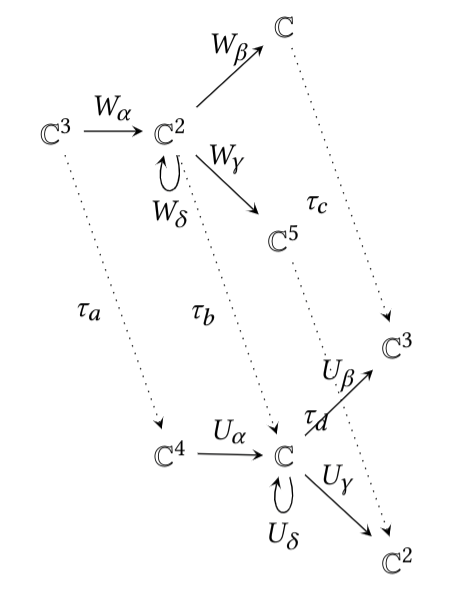}
%  \caption{Diagram of a morphism of representations.}
%\end{subfigure}%
%\begin{subfigure}{.4\textwidth}
%  \centering
(b)  \includegraphics[width=.4\linewidth]{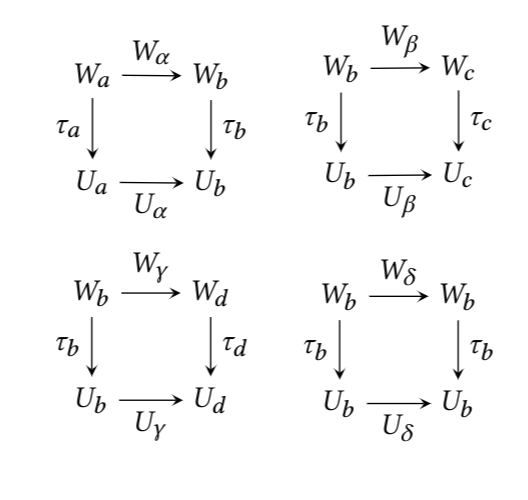}
%  \caption{Diagrams that should be commutative for $\tau$ to be a morphism of representations.}
%\end{subfigure}%
\caption{(a) A morphism of representations $\tau:W \to U$ is given by a family of matrices $\tau = (\tau_v)_{v \in \V}$, such that $\tau_a: \Cmplx^3 \to \Cmplx^4$, $\tau_b: \Cmplx^2 \to \Cmplx$, $\tau_c: \Cmplx \to \Cmplx^3$ and $\tau_d:\Cmplx^5 \to \Cmplx^2$ satisfy that $\tau_b W_\alpha = U_\alpha \tau_a,  \tau_c W_\beta = U_\beta \tau_b, \tau_d W_\gamma = U_\gamma \tau_b,  \tau_b W_\delta = U_\delta \tau_b$. (b) Four diagrams showing that the transformations $\tau_v$ must make them commutative for $\tau:W \to U$ to be a morphism of representations.} \label{fig:morphism_reps}
\end{figure}

\begin{Definition}  \citep[chap.~3]{Assem06}
    Let $Q$ be a quiver and let $W$ and $U$ be two representations of $Q$. A \textbf{morphism of representations} $\tau : W \to U$ is a set of linear maps $\tau = (\tau_v)_{v\in \V}$ indexed by the vertices of $Q$, where $\tau_v:W_v \to U_v$ is a linear map such that $\tau_{t(\epsilon)} W_\epsilon = U_\epsilon \tau_{s(\epsilon)}$ for every $\epsilon \in \E$.
\end{Definition}
To illustrate this definition, one may consider the quiver $Q$ and its representations $W$ and $U$ of Fig.~\ref{fig:Quiver_Reps}.
The morphism between $W$ and $U$ via the linear maps $\tau$ are pictured in Fig.~\ref{fig:morphism_reps}(a).  As shown, each $\tau_v$ is a matrix which allows to transform the vector space of vertex $v$ of $W$ into the vector space of vertex $v$ of $U$.

\begin{Definition}  
    Let $Q$ be a quiver and let $W$ and $U$ be two representations of $Q$. If there is a morphism of representations $\tau : W \to U$ where each $\tau_v$ is an invertible linear map, then $W$ and $U$ are said to be \textbf{isomorphic representations}.
\end{Definition}

{\color{black}
The previous definition is equivalent to the usual categorical definition of isomorphism, see \cite[chap.3~]{Assem06}. Namely, a morphism of representations $\tau:W \to U$ is an isomorphism if there exists a morphism of representations $\eta : U \to W$ such that $\eta \circ \tau = id_W$ and $\tau \circ \eta = id_U$. Observe here that the composition of morphisms is defined as a coordinate wise composition, indexed by the vertices of the quiver.
}

In Section~\ref{sec:neuralNets}, we will be working with a particular type of quiver representations, where the vector space of each vertex is in 1D. This 1D representations are called thin representations, and the morphisms of representations between thin representations are easily described.

{\color{black}
\begin{Definition}
    A \textbf{thin representation} of a quiver $Q$ is a quiver representation $W$ such that $W_v=\Cmplx$ for all $v \in V$.
\end{Definition}
}

%\begin{Remark}
    If $W$ is a thin representation of $Q$, then every linear map $W_\epsilon$ is a $1 \times 1$ matrix, so $W_\epsilon$ is given by multiplication with a fixed complex number. We may and will identify every linear map between one dimensional spaces with the number whose multiplication defines it.
%\end{Remark}

Before we move on to neural networks, we will introduce the notion of group and action of a group.
\begin{Definition} \citep[chap.~1]{Rotman95}
    A non-empty set $G$ is called a \textbf{group} if there exists a function %\newline 
    $\cdot : G \times G \to G$, called the product of the group denoted $a \cdot b$, such that
    \begin{itemize}
        \item $(a \cdot b) \cdot c = a \cdot (b \cdot c)$ for all $a,b,c \in G$.
        \item There exists an element $e \in G$ such that $e\cdot a = a \cdot e = a$ for all $a\in G$, called the \textbf{identity} of $G$.
        \item For each $a\in G$ there exists $a^{-1} \in G$ such that $a \cdot a^{-1} = a^{-1} \cdot a = e$.
    \end{itemize}
\end{Definition}
For example, the set of non-zero complex numbers $\mathbb{C}^*$ (and also the non-zero real numbers $\mathbb{R}^*$) with the usual multiplication operation forms a group. Usually, one does not write the product of the group as a dot and just concatenates the elements to denote multiplication $ab = a\cdot b$, as for the product of numbers.
\begin{Definition} \citep[chap.~3]{Rotman95} \label{def:actiongroup}
    Let $G$ be a group and let $X$ be a set. We say that there is an \textbf{action of} G \textbf{on} X if there exists a map $\cdot: G \times X \to X$ such that
    \begin{itemize}
        \item $e \cdot x = x$ for all $x\in X$, where $e\in G$ is the identity.
        \item $a \cdot (b \cdot x) = (a b) \cdot x$, for all $a,b \in G$ and all $x\in X$.
    \end{itemize}
\end{Definition}

In our case, $G$ will be a group indexed by the vertices of $Q$, and the set $X$ will be the set of thin quiver representations of $Q$.

Let $W$ be a thin representation of a quiver $Q$. Given a choice of invertible (non-zero) linear maps $\tau_v: \Cmplx \to \Cmplx$ for every $v\in \V$, we are going to construct a thin representation $U$ such that $\tau=(\tau_v)_{v \in \V} :W \to U$ is an isomorphism of representations. Since $U$ is thin, we have that $U_v=\Cmplx$ for all $v\in \V$. Let $\epsilon:a \to b$ be an edge of $\E$, we define the group action as follows,
\begin{eqnarray}
\label{eq:group_action}
U_\epsilon = W_\epsilon \cdot \dfrac{\tau_b}{\tau_a}.
\end{eqnarray}
Thus, for every edge $\epsilon \in \mathcal{E}$ we get a commutative diagram
\[
    \begin{tikzpicture}
     	\matrix (m) [matrix of math nodes,row sep=2em,column sep=2em]
      	{
       	  	W_{s(\epsilon)} & W_{t(\epsilon)}   \\
       	  	U_{s(\epsilon)} & U_{t(\epsilon)}.   \\
       	};
      	\path[-stealth]
      	 (m-1-1) edge node [above] {$W_\epsilon$} (m-1-2)
       	 (m-2-1) edge node [below] {$U_\epsilon$} (m-2-2)
       	 (m-1-1) edge node [left] {$\tau_{s(\epsilon)}$}  (m-2-1)
       	 (m-1-2) edge node [right] {$\tau_{t(\epsilon)}$} (m-2-2);
    \end{tikzpicture}
\]
The construction of the thin representation $U$ from the thin representation $W$ and the choice of invertible linear maps $\tau$, defines an action on thin representations of a group. The set of all possible isomorphisms $\tau = (\tau_v)_{v\in \V}$ of thin representations of $Q$ forms such a group.

{\color{black}
\begin{Definition}
    The \textbf{change of basis group} of thin representations over a quiver $Q$ is
    \[
    G = \displaystyle\prod_{v \in \V} \Cmplx^*,
    \]
    where $\Cmplx^*$ denotes the multiplicative group of non-zero complex numbers. That is, the elements of $G$ are vectors of non-zero complex numbers $\tau = (\tau_1,...,\tau_n)$ indexed by the set $\V$ of vertices of $Q$, and the group operation between two elements $\tau = (\tau_1,...,\tau_n)$ and $\sigma = (\sigma_1,...,\sigma_n)$ is by definition
    \[
        \tau \sigma := (\tau_1 \sigma_1,..., \tau_n \sigma_n).
    \]  
\end{Definition}
}
%called the \textbf{change of basis group} $G$ defined as the product group

We use the action notation for the action of the group $G$ on thin representations. Namely, for $\tau \in G$ of the form $\tau = (\tau_v)_{v\in \V}$ and a thin representation $W$ of $Q$, the thin representation $U$ constructed above is denoted $\tau \cdot W$.

\section{Neural Networks}
\label{sec:neuralNets}

%In this section, we will introduce the mathematical objects that will contain the structure of a neural network using ideas, constructions and results from quiver representation theory. 
%It is usual in mathematics \citep{Armenta19} and physics \citep{Einstein50} to extend the domains of a problem in order to develop theory with which we understand the simpler case. In our case, we will extend neural networks to the complex numbers. However, we should not forget that in practice we will have very particular complex numbers, those that have zero imaginary part, i.e., real numbers. Although this approach may seem strange, the use of complex numbers expands the amount of algebra we can use to study neural networks \citep{Assem06,Reineke08}. Even more, we will show how the mathematical consequences of using complex numbers provide strong connections with the current understanding of deep learning. Note also the existence of complex back propagation \citep{Nitta97}.
In this section, we connect the dots between neural networks and the basic definitions of quiver representation theory that we presented before.
But before we do so, let us mention that since the vector space of each vertex of a quiver representation is defined over the complex numbers, it implies that the weights on the neural networks that we are to present will also be complex numbers.  Despite some papers on complex neural networks~\citep{Nitta97}, this approach may seem unorthodox.  However, the use of complex numbers is a mathematical pre-requisite for the upcoming notion of moduli space that we will introduce in Section~\ref{sec:moduli}. {\color{black} Observe also, that this does not mean that in practice neural networks should be based on complex numbers. It only means that neural networks in practice, which are based upon real numbers, trivially satisfy the condition of being complex neural networks, and therefore the mathematics derived from using complex numbers apply to neural networks over real numbers.}

%provide strong connections with the current understanding of deep learning.  % We will denote by $\Cmplx^n$ the complex vector space of dimension $n$.

For the rest of this paper, we will focus on a special type of quiver $Q$ that we call \textit{network quiver}.  A network quiver $Q$ has no oriented cycles other than loops. Also, a sub-set of $d$ source vertices of $Q$ are called the \textbf{input vertices}. The source vertices that are not input vertices are called \textbf{bias vertices}.  Let $k$ be the number of all sinks of $Q$, we call these the \textbf{output vertices}.  All other vertices of $Q$ are called \textbf{hidden vertices}.
\begin{Definition}
    A quiver $Q$ is \textbf{arranged by layers} if it can be drawn from left to right arranging its vertices in columns such that:
    \begin{itemize}
        \item There are no oriented edges from vertices on the right to vertices on the left.
        \item There are no oriented edges between vertices in the same column, other than loops and edges from bias vertices.
    \end{itemize}
    The first layer on the left, called the \textbf{input layer}, will be formed by the $d$ input vertices. The last layer on the right, called the \textbf{output layer}, will be formed by the $k$ output vertices. The layers that are not input nor output layers are called \textbf{hidden layers}.  We enumerate the hidden layers from left to right as : $1^{\mbox{\scriptsize st}}$ hidden layer, $2^{\mbox{\scriptsize nd}}$ hidden layer, $3^{\mbox{\scriptsize rd}}$ hidden layer, and so on.
\end{Definition}

\begin{figure} 
%\centering
%\begin{subfigure}{.5\textwidth}
%  \centering
(a)  \includegraphics[width=.35\linewidth]{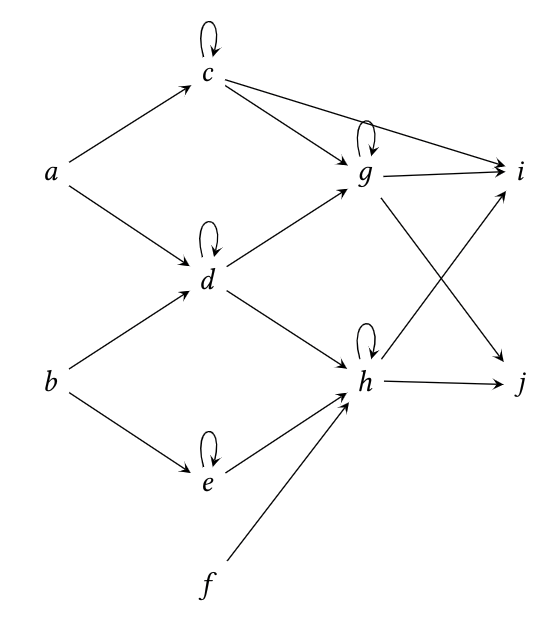}
%  \caption{A network quiver.}
%\end{subfigure}%
%\begin{subfigure}{.5\textwidth}
%  \centering
(b)  \includegraphics[width=.4\linewidth]{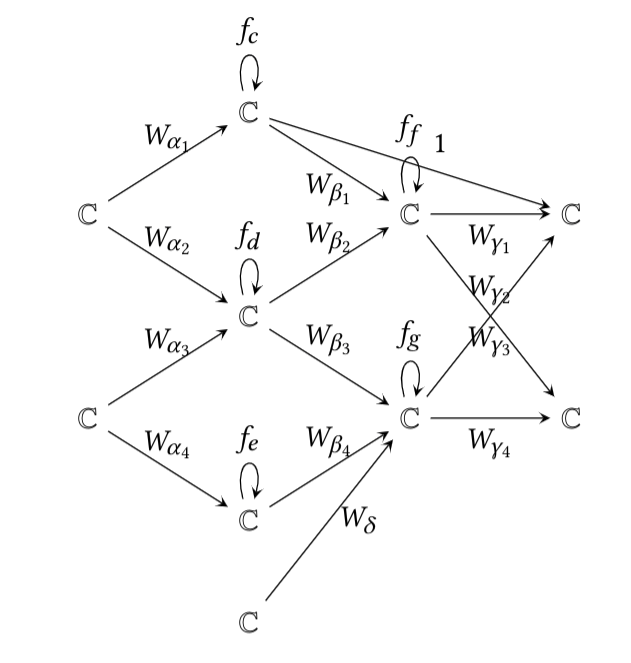}
%  \caption{A neural network }
%\end{subfigure}
\caption{(a) A network quiver $Q$ whose input layer is given by the vertices $a$ and $b$, the vertex $f$ is a bias vertex and there is a skip connection from vertex $c$ to vertex $i$.  Note that we did not label the edges to lighten the diagram.  (b) A neural network over $Q$ where $W_{\alpha_1}, W_{\alpha_2} ..., W_{\delta}$ are linear maps given by multiplication with a number, and the functions $f=(f_c,f_d,f_e,f_f,f_g)$ are the activation functions (could be sigmoid, tanh, ReLU, ELU, etc.)} \label{fig:Net_Quiv_Ex}
\end{figure}

From now on $Q$ will always denote a quiver with $d$ input vertices and $k$ output vertices.
\begin{Definition}
    A \textbf{network quiver} $Q$ is a quiver arranged by layers such that:
    \begin{itemize}
        \item[1.] There are no loops on source (i.e., input and bias) nor sink vertices.
        \item[2.] There is exactly one loop on each hidden vertex.
        %\item[3.] Other than these loops, there are no more oriented cycles.
    \end{itemize}
\end{Definition}
An example of a network quiver can be found in Fig.~\ref{fig:Net_Quiv_Ex}(a).
\begin{Definition}
    The \textbf{delooped} quiver $Q^\circ$ of $Q$ is the quiver obtained by removing all loops of $Q$. We denote $Q^\circ = (\V, \mathcal{E}^\circ, s^\circ, t^\circ)$.
\end{Definition}

    When a neural network computes a forward pass (be it a multilayer perceptron, a convolutional neural network {\color{black} and even a randomly wired neural network \cite{Xie19}}), the weight between two neurons is used to multiply the output signal of the first neuron and the result is fed to the second neuron. Since multiplying {\color{black}with a number (the weight)} is a linear map, we get that a weight is used as a linear map between two 1D vector spaces {\color{black}during inference}. Therefore the weights of a neural network define a thin quiver representation of the delooped quiver $Q^\circ$ of its network quiver $Q$, every time it computes a prediction. 
    
    When a neural network computes a forward pass, we get a combination of two things:
    \begin{enumerate}
        \item[1.] A thin quiver representation.
        \item[2.] Activation functions.
    \end{enumerate}
    {\color{black}
    \begin{Definition}
        An \textbf{activation function} is a one variable non-linear function $f: \Cmplx \to \Cmplx$ differentiable except in a set of measure zero.
    \end{Definition}
    }
    
    {\color{black}
    \begin{Remark}
        An activation function can, in principle, be linear. Nevertheless, neural network learning occurs with all its benefits only in the case where activation functions are fundamentally non-linear. Here, we want to provide a universal language for neural networks, so we will work with neural networks with non-linear activation functions, unless explicitly stated otherwise, for example as in our data representations in Section~\ref{sec:datarep}.
    \end{Remark}
    }
    
    We will encode the point-wise usage of activation functions as maps assigned to the loops of a network quiver. 
    
%\end{Remark}
\begin{Definition}
    A \textbf{neural network} over a network quiver $Q$ is a pair $(W,f)$ where $W$ is a thin representation of the delooped quiver $Q^\circ$ and $f=(f_v)_{v \in \V}$ are activation functions, assigned to the loops of $Q$.
\end{Definition}
An example of neural network $(W,f)$ over a network quiver $Q$ can be seen in Fig.~\ref{fig:Net_Quiv_Ex}(b).  The words \textbf{neuron} and \textbf{unit} refer to the combinatorics of a vertex together with its activation function in a neural network over a network quiver. The \textbf{weights} of a neural network $(W,f)$ are the complex numbers defining the maps $W_\epsilon$ for all $\epsilon \in \E$.
%\end{Remark}

When computing a prediction, we have to take into account two things:
    \begin{itemize}
        \item The activation function is applied to the sum of all input values of the neuron.
        \item The activation output of each vertex is multiplied by each weight going out of that neuron.
    \end{itemize}

{\color{black}
Once a network quiver and a neural network $(W,f)$ are chosen, a decision has to be made on how to compute with the network. For example, a hidden neuron may compute an inner product of its inputs followed by the activation function, but others, like max-pooling, output the maximum of the input values. We account for this by specifying in the next definition how every type of vertex is used to compute. 
}

\begin{Definition} \label{def:actfun}
    Let $(W,f)$ be a neural network over a network quiver $Q$ and let $x \in \Cmplx^d$ be an input vector of the network. Denote by $\zeta_v$ the set of edges of $Q$ with target $v$. The \textbf{activation output of the vertex} $v \in \V$ \textbf{with respect to} $x$ after applying a forward pass is denoted $\act(W,f)_v(x)$ and is computed as follows:
    \begin{itemize}
        \item If $v\in \V$ is an input vertex, then $\act(W,f)_v(x) = x_v$.
        \item If $v \in \V$ is a bias vertex, then $\act(W,f)_v(x)=1$.
        \item If $v \in \V$ is a hidden vertex, then $\act(W,f)_v(x) = f_v \left( \displaystyle\sum_{\alpha \in \zeta_v} W_\alpha \act(W,f)_{s(\alpha)}(x)  \right)$.
        \item If $v\in \V$ is an output vertex, then $\act(W,f)_v(x) = \displaystyle\sum_{\alpha \in \zeta_v} W_\alpha \act(W,f)_{s(\alpha)} (x)$.
        {\color{black}
        \item If $v\in \V$ is a max-pooling vertex, then $\act(W,f)_v(x) = \maxx_\alpha Re\big( W_{\alpha} \act(W,f)_{s(\alpha)}(x)  \big)$, where $Re$ denotes the real part of a complex number, and the maximum is taken over all $\alpha\in \mathcal{E}$ such that $t(\alpha)=v$.
        }
    \end{itemize}
\end{Definition}

{\color{black} We will see in the next chapter how and why average pooling vertices do not require a different specification on the computation rule, because it can be written in terms of these same rules.

The previous definition is equivalent to the basic operations of a neural net, which are affine transformations followed by point-wise non-linear activation functions, see Appendix~\ref{appendix1} where we clarify this with an example. The advantage of using the combinatorial expression of Definition \ref{def:actfun} is twofold, (i) it allows to represent any architecture, even randomly wired neural networks \citep{Xie19}, and (ii) it allows to simplify the notation on proofs concerning the network function.}

For our purposes, it is convenient to consider no activation functions on the output vertices. This is consistent with current deep learning practices as one can consider the activation functions of the output neurons to be part of the loss function (like softmax + cross-entropy or as done by~\citet{Dinh17}).

\begin{Definition}
    Let $(W,f)$ be a neural network over a network quiver $Q$. The \textbf{network function} of the neural network is the function 
    \[
        \Psi(W,f):\Cmplx^d \to \Cmplx^k
    \]
    where the coordinates of $\Psi(W,f)(x)$ are the activation outputs of the output vertices of $(W,f)$ (often called the ``score" of the neural net) with respect to an input vector $x \in \Cmplx^d$.
\end{Definition}
%\begin{Remark}
    
%\end{Remark}
{ \color{black} 
The only difference in our approach is the combinatorial expression of Definition~\ref{def:actfun} which can be seen as a neuron-wise computation, that in practice is performed by layers {\color{black} for implementation purposes}. These expressions will be useful to prove our more general results.

We now extend the notion of isomorphism of quiver representations to isomorphism of neural networks. For this, we have to take into account that isomorphisms of quiver representations carry the commutative diagram conditions given by all the edges in the quiver, as shown in Fig.~\ref{fig:morphism_reps}. For neural networks, the activation functions are non-linear, but this does not prevents us from putting a commutative diagram condition on activation functions as well. So an isomorphism of quiver representations acts on a neural network in the sense of the following definition.
}
%We now define maps that behave well with respect to the structure of neural networks, namely the thin quiver representation and the activation functions at the same time.
\begin{Definition}
    Let $(W,f)$ and $(V,g)$ be neural networks over the same network quiver $Q$. A \textbf{morphism of neural networks} $\tau:(W,f) \to (V,g)$ is a morphism of thin quiver representations $\tau:W \to V$ such that $\tau_v=1$ for all $v \in \V$ that is not a hidden vertex, and for every hidden vertex $v \in \V$ the following diagram is commutative
    \[
        \begin{tikzpicture}
         	\matrix (m) [matrix of math nodes,row sep=2em,column sep=2em]
          	{
           	  	\Cmplx  & \Cmplx \\
           	  	\Cmplx & \Cmplx.  \\
           	};
          	\path[-stealth]
           	 (m-1-1) edge node [above] {$f_v$} (m-1-2)
           	 (m-1-2) edge node [right] {$\tau_v$} (m-2-2)
            
           	 (m-1-1) edge node [left] {$\tau_v$} (m-2-1)
           	 (m-2-1) edge node [below] {$g_v$} (m-2-2);
	    \end{tikzpicture}
    \]
    A morphism of neural networks $\tau:(W,f) \to (V,g)$ is an \textbf{isomorphism of neural networks} if $\tau:W \to V$ is an isomorphism of quiver representations.
    We say that two neural networks over $Q$ are \textbf{isomorphic} if there exists an isomorphism of neural networks between them.
\end{Definition}
%\begin{Remark}
    %Every morphism of neural networks is an isomorphism of neural networks due to the condition that the change of basis is equal to the identity on the source and output vertices.
%\end{Remark}
\begin{Remark}
    The terms 'network morphism'~\citep{Wei16}, 'isomorphic neural network' and 'isomorphic network structures'~\citep{Meng19,Stagge00} have already been used with different approaches. In this work, we will not refer to any of those terms.
\end{Remark}

{ \color{black} 
\begin{Definition}
     The \textbf{hidden quiver} of $Q$, denoted by $\Q = (\VV, \widetilde{\mathcal{E}}, \widetilde{s}, \widetilde{t})$, is given by the hidden vertices $\VV$ of $Q$ and all the oriented edges $\widetilde{\mathcal{E}}$ between hidden vertices of $Q$ that are not loops.
\end{Definition}
}

 Said otherwise, $\Q$ is the same as the delooped quiver $Q^\circ$ but without the source and sink vertices.  
 
{\color{black}
\begin{Definition}
     The \textbf{group of change of basis} for neural networks is denoted as
\[
    \Gt = \displaystyle\prod_{v \in \VV} \Cmplx^*.
\]
An element of the change of basis group $\Gt$ is called a \textbf{change of basis} of the neural network $(W,f)$.
\end{Definition}
}

Note that this group has as many factors as hidden vertices of $Q$. 
%\begin{Remark}
    Given an element $\widetilde{\tau} \in \Gt$ we can induce $\tau \in G$, where $G$ is the change of basis group of thin representations over the delooped quiver $Q^\circ$. We do this by assigning $\tau_v=1$ for every $v \in \V$ that is not a hidden vertex. Therefore, we will simply write $\tau$ for elements of $\Gt$ considered as elements of $G$.
%\end{Remark}

The action of the group $\Gt$ on a neural network $(W,f)$ is defined on a given element $\tau \in \Gt$ and a neural network $(W,f)$ by
\[
    \tau \cdot (W,f) = (\tau \cdot W, \tau \cdot f),
\]
where $\tau \cdot W$ is the thin representation such that for each edge $\epsilon \in \E$, the linear map $\big( \tau \cdot W \big)_\epsilon = W_\epsilon \dfrac{\tau_{t(\epsilon)}}{\tau_{s(\epsilon)}}$ following the group action of Eq.(\ref{eq:group_action}), and the activation $\tau \cdot f$ on the hidden vertex $v \in \V$ is given by 
\begin{eqnarray}
\label{eq:activation_action}
(\tau \cdot f)_v(x)=\tau_v f \left( \dfrac{x}{\tau_v} \right) \mbox{ for all } x\in \Cmplx.  
\end{eqnarray}

Observe that $(\tau \cdot W, \tau \cdot f)$ is a neural network such that $\tau : (W,f) \to (\tau \cdot W, \tau \cdot f) $ is an isomorphism of neural networks.  This leads us to the following theorem, which is an important corner stone of our paper.  Please refer to Appendix~\ref{appendix1} for an illustration of this proof.

\begin{Theorem} \label{thm:netfunc}
    If $\tau:(W,f) \to (V,g)$ is an isomorphism of neural networks, then $\Psi(W,f)=\Psi(V,g).$
\end{Theorem}
 {\bf Proof}.
    Let $\tau:(W,f) \to (V,g)$ be an isomorphism of neural networks over $Q$ and $\epsilon:s(\epsilon) \to t(\epsilon)$ an oriented edge of $Q$.  Considering the group action of Eq.(\ref{eq:group_action}), if $s(\epsilon)$ and $t(\epsilon)$ are hidden vertices then $V_\epsilon=W_\epsilon \cdot \dfrac{\tau_{t(\epsilon)}}{\tau_{s(\epsilon)}}$.
    However, if $s(\epsilon)$ is a source vertex, then $\tau_{s(\epsilon)}=1$ and $V_\epsilon=W_\epsilon \tau_{t(\epsilon)}$. And if $t(\epsilon)$ is an output vertex, then $\tau_{t(\epsilon)}=1$ and $V_\epsilon = \dfrac{W_\epsilon}{\tau_{s(\epsilon)}}$. Also, for every hidden vertex $v\in \VV$ we get the activation function $g_v(z)=\tau_v \cdot f_v\left( \dfrac{z}{\tau_v} \right)$ for all $z \in \Cmplx$. 
    
    We proceed with a forward pass to compare the activation outputs of both neural networks with respect to the same input vector. Let $x \in \Cmplx^d$ be the input vector of the networks, for every source vertex $v\in \V$ we have
    \begin{eqnarray}
        \act(W,f)_v(x) = \act(V,g)_v(x) = \left \{
        \begin{array}{lr}
        x_v \in \Cmplx & \mbox{ if } v \mbox{ is an input neuron,}\\
        1 & \mbox{if } v \mbox{ is a bias neuron.}
        \end{array}
        \right .
        \label{eq:fp_activation_ouput}
    \end{eqnarray}
    Now let $v\in \V$ be a vertex in the first hidden layer and $\zeta_v$ the set of edges between the source vertices and $v\in \V$, the activation output of $v$ in $(W,f)$ is
    \begin{eqnarray}
        \act(W,f)_v(x) = f_v \left( \displaystyle\sum_{\epsilon \in \zeta_v} W_\epsilon \cdot \act(W,f)_{s(\epsilon)}(x) \right). \nonumber
    \end{eqnarray}
    As an illustration, if $(W,f)$ is the neural network of Fig.~\ref{fig:Net_Quiv_Ex}, the source vertices would be $a,b,f$, the first hidden layer vertices would be $c,d,e$ and the weights $W_\epsilon$ in the previous equation would be $\{ W_{\alpha_2}, W_{\alpha_3}\}$ when $v=d$.
    We  now calculate in $(V,g)$ the activation output of the same vertex $v$,
    \[
        \begin{array}{lcl}
            \act(V,g)_v(x) & = & \tau_v f_v \left( \dfrac{1}{\tau_v} \displaystyle\sum_{\epsilon \in \zeta_v} V_\epsilon \cdot \act(V,g)_{s(\epsilon)}(x) \right) \\
            & = & \tau_v f_v \left( \dfrac{1}{\tau_v} \displaystyle\sum_{\epsilon \in \zeta_v} W_\epsilon \tau_{t(\epsilon)} \cdot \act(V,g)_{s(\epsilon)}(x) \right)  
        \end{array}
    \]
    since $t(\epsilon)=v$, then $\tau_{t(\epsilon)}=\tau_v$ and
    \[
             = \tau_v f_v \left( \displaystyle\sum_{\epsilon \in \zeta_v}
            W_\epsilon \cdot \act(V,g)_{s(\epsilon)}(x) \right) 
    \]
    and since $s(\epsilon)$ is a source vertex, it follows from Eq.(\ref{eq:fp_activation_ouput}) that $\act(W,f)_{s(\epsilon)}(x)=\act(V,g)_{s(\epsilon)}(x)$ and
    \[
        \begin{array}{lcl}
            & = & \tau_v f_v \left( \displaystyle\sum_{\epsilon \in \zeta_v} W_\epsilon \cdot \act(W,f)_{s(\epsilon)}(x) \right)  \\
            & = & \tau_v \act(W,f)_v(x),
        \end{array}
    \]

 Assume now that $v\in \V$ is in the second hidden layer (e.g., vertex $g$ or $h$ in Fig.~\ref{fig:Net_Quiv_Ex}), the activation output of $v$ in $(V,g)$ is
    \[
        \begin{array}{lcl}
            \act(V,g)_v(x) & = & \tau_v f_v \left( \dfrac{1}{\tau_v}  \displaystyle\sum_{\epsilon \in \zeta_v} V_\epsilon \cdot \act(V,g)_{s(\epsilon)}(x) \right) \\
             & = & \tau_v f_v \left( \dfrac{1}{\tau_v}  \displaystyle\sum_{\epsilon \in \zeta_v}  \dfrac{W_\epsilon \tau_v }{\tau_{s(\epsilon)}} \act(V,g)_{s(\epsilon)}(x) \right) \\
             & = & \tau_v f_v \left( \displaystyle\sum_{\epsilon \in \zeta_v}  \dfrac{W_\epsilon }{\tau_{s(\epsilon)}} \act(V,g)_{s(\epsilon)}(x) \right)
        \end{array}
\] 
and since $\act(V,g)_{s(\epsilon)}(x) = \tau_{s(\epsilon)} \act(W,f)_{s(\epsilon)}(x)$ from the equation above, then
\[
     \begin{array}{lcl}
             & = & \tau_v f_v \left( \displaystyle\sum_{\epsilon \in \zeta_v}  \dfrac{W_\epsilon }{\tau_{s(\epsilon)}} \tau_{s(\epsilon)} \act(W,f)_{s(\epsilon)}(x) \right) \\
             & = & \tau_v f_v \left( \displaystyle\sum_{\epsilon \in \zeta_v}  W_\epsilon  \act(W,f)_{s(\epsilon)}(x) \right) \\
             & = & \tau_v \act(W,f)_v(x).
        \end{array}
    \]
    Inductively, we get that $\act(V,g)_v(x) = \tau_v \act(W,f)_v(x)$ for every vertex $v\in \V$. Finally, the coordinates of $\Psi(W,f)(x)$ are the activation outputs of $(W,f)$ on the output vertices, and analogously for $\Psi(V,g)(x)$. Since $\tau_v=1$ for every output vertex $v\in \V$, we get that
    \[
        \Psi(W,f)(x) = \Psi(V,g)(x)
    \]
    which proves that an isomorphism between two neural networks $(W,f)$ and $(V,g)$ preserves the network function.\hfill\BlackBox
 
{\color{black}
\begin{Remark}
    Max-pooling represents a different operation to obtain the activation output of neurons. After applying an isomorphism $\tau$ to a neural network $(W,f)$, where the vertex $v\in \V$ is a max-pooling vertex we obtain an isomorphic neural network $(V,g)$, whose activation output on vertex $v$ is given by the following formula:
    \[
    \act(V,g)_v(x) = \left\{ \begin{array}{ll}
         {\displaystyle\maxx_{\begin{array}{l}
              \alpha \in \mathcal{E} \\ t(\alpha)=v
        \end{array}}} Re\big( V_{\alpha} \act(V,g)_{s(\alpha)}(x)  \big) & \text{ if } Re(\tau_v) \geq 0 \\ \\
        {\displaystyle\minn_{\begin{array}{l}
              \alpha \in \mathcal{E} \\ t(\alpha)=v
        \end{array}}} Re\big( V_{\alpha} \act(V,g)_{s(\alpha)}(x)  \big) & \text{ if } Re(\tau_v) < 0,
    \end{array} \right.
    \]
    and 
    \[
        V_\alpha \act(V,g)_{s(\alpha)}(x) = \tau_{t(\alpha)} W_\alpha \tau_{s(\alpha)}^{-1} \tau_{s(\alpha)} \act(W,f)_{s(\alpha)}(x) = \tau_{t(\alpha)} W_\alpha \act(W,f)_{s(\alpha)}(x),
    \]
    which is the main argument in the proof of the previous theorem, so the result applies to max-pooling. Note also that max-pooling vertices are positive scale invariant.
\end{Remark}
}

\subsection{Consequences}

Representing a neural network over a network quiver $Q$ by a pair $(W,f)$ and Theorem~\ref{thm:netfunc} has two consequences on neural networks.

\paragraph{Consequence 1}
{\color{black}
\begin{Corollary}
    There are infinitely many neural networks with the same network function, independently of the architecture and the activation functions.
\end{Corollary}
}

If each neuron of a neural network is assigned a change of basis value $\tau_v \in \Cmplx$, its weights $W$ can be transformed to another set of weights $V$ following the group action of Eq.(\ref{eq:group_action}).  Similarly, the activation functions $f$ of that network can be transformed to other ones $g$ following the group action of Eq.(\ref{eq:activation_action}).  For example, if $f$ is ReLU and $\tau_v$ is a negative real value, then $g$ becomes an inverted-flipped ReLU function, i.e., $min(0,x)$.  From the usual neural network representation stand point, %the one which represents a neural network by its architecture or as a point in a N dimensional space when N is the number of weights lying on the slope of a loss landscape, 
the two neural networks $(W,f)$  and $(V,g)$ are different as their activation functions $f$ and $g$ are different and their weights $W$ and $V$ are different.  Nonetheless, their function (i.e., the output of the networks given some input vector $x$) is rigorously identical.  And this is true regardless of the structure of the neural network, its activation functions and weight vector $W$.

Said otherwise, Theorem~\ref{thm:netfunc} implies that there is not a unique neural network with a given network function and that an [infinite] amount of other neural networks with different weights and different activation functions have the same network function and that these other neural networks may be obtained with the change of basis group $\Gt$.

\paragraph{Consequence 2} A weak version of Theorem~\ref{thm:netfunc} proves a property of ReLU networks known as positive scale invariance or positive homogeneity~\citep{Badrinarayanan15,Dinh17,Meng18,Yi19,Yuan19}. Positive scale invariance is a property of ReLU non-linearities, where the network function remains unchanged if we (for example) multiply the weights in one layer of a network by a positive factor, and divide the weights on the next layer by that same positive factor. Even more, this can be done on a per neuron basis. Namely, assigning a positive factor $r>0$ to a neuron and multiplying every weight that points to that neuron with $r$, and dividing every weight that starts on that neuron by $r$.

\begin{Corollary} (Positive Scale Invariance of ReLU Networks) \label{cor:pos_sca_inv}
    Let $(W,f)$ be a neural network over $Q$ over the real numbers where $f$ is the ReLU activation function. Let $\tau=(\tau_v)_{v \in \V}$ where $\tau_v=1$ if $v$ is not a hidden vertex, and $\tau_v>0$ for any other $v$. Then 
    \[
        \tau \cdot (W,f) = (\tau \cdot W, f).
    \]
    As a consequence, $(\tau \cdot W,f)$ and $(W,f)$ are isomorphic neural networks. In particular, they have the same network function, $\Psi(\tau \cdot W,f) = \Psi(W,f)$.
\end{Corollary}
 {\bf Proof}.
    Recall that $\tau \cdot (W,f) = (\tau \cdot W, \tau \cdot f)$. Since ReLU satisfies $f(\tau_v x) = \tau_v f(x)$ for all $x$ and all $\tau_v>0$ and since $(\tau \cdot f)$ corresponds to $ \tau_v f \left( \dfrac{x}{\tau_v} \right)$ at each vertex $v$ as mentioned in Eq.(\ref{eq:activation_action}), we get that $ \tau_v f \left( \dfrac{x}{\tau_v} \right)=\dfrac{\tau_v}{\tau_v} f \left( x \right) = f \left( x \right)$ for each vertex $v$ and thus $\tau \cdot f = f$. Finally, $\tau \cdot(W,f) = (\tau \cdot W, \tau \cdot f) = (\tau \cdot W, f)$.\hfill\BlackBox

{\color{black}We stress out that this known result is a consequence of neural networks being pairs $(W,f)$ whose structure is governed by representation theory, and therefore exposes the algebraic and combinatorial nature of neural networks.}

\section{Architecture} \label{sec:architecture}

In this section, we first outline the different types of architectures that we consider. We also show how the commonly used layers for neural networks translate into quiver representations. Finally, we will present in detail how an isomorphism of neural networks can be chosen so that the structure of the weights gets preserved.

\subsection{Types of architectures}

\begin{Definition} \citep[page~193]{Goodfellow-et-al-2016}
    The \textbf{architecture} of a neural network refers to its structure which accounts for how many units (neurons) it has and how these units are connected together.
\end{Definition}
For our purposes, we distinguish three types of architectures: \textbf{combinatorial architecture}, \textbf{weight architecture} and \textbf{activation architecture}.
\begin{Definition}
    The \textbf{combinatorial architecture} of a neural network {\color{black}is its} network quiver. The \textbf{weight architecture} is given by constraints on how the weights are chosen, and the \textbf{activation architecture} is the set of activation functions assigned to the loops of the network quiver.
\end{Definition}
%Observe that the combinatorial architecture encodes the sequence of operations that a neural network computes. The weight architecture encodes the linear maps given by the weights and the activation architecture encodes how the activations are applied. 

If we consider the neural network of Fig.~\ref{fig:Net_Quiv_Ex}, the combinatorial architecture specifies how the vertices are connected together, the weight architecture on how the weights $W_\epsilon$ are assigned and the activation architecture deals with the activation functions $f_v$. 

Two neural networks may have different combinatorial, weight and activation architecture like ResNet~\citep{He16} vs VGGnet~\citep{Simonyan15} for example.  Neural network layers may have the same combinatorial architecture but a different activation and weight architecture.  It is the case for example of a mean pooling layer vs a convolution layer.  While they both encode a convolution (same combinatorial architecture) they have a different activation architecture (as opposed to conv layers, mean pooling has no activation function) and a different weight architecture as the mean pooling weights are fixed, {\color{black}and on conv layers they are shared across filters. This is what we mean by ``\textit{constraints}'' on how the weights are chosen, namely, weights in conv layers and mean-pooling layers are not chosen freely, as in fully connected layers.}  Overall, two neural networks have globally the same architecture if and only if they share the same combinatorial, weight, and activation architectures.  

Also, isomorphic neural networks always have the same combinatorial architecture, since isomorphisms of neural networks are defined over the same network quiver. However, an isomorphism of neural networks can change or not the weight and the activation architecture. We will come back on that concept at the end of this section.

\subsection{Neural network layers}
Here, we look at how fully-connected layers, convolutional layers, pooling layers, batch normalization layers and residual connections are related to the quiver representation language.

\begin{figure} 
%\centering
%\begin{subfigure}{.5\textwidth}
%  \centering
\begin{center}
    (a)  \includegraphics[width=.16\linewidth]{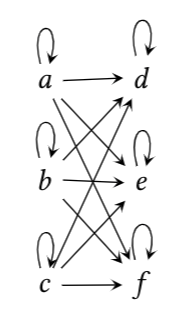}
%  \caption{A fully connected layer.}
%\end{subfigure}%
%\begin{subfigure}{.5\textwidth}
%  \centering
(b)  \includegraphics[width=.25\linewidth]{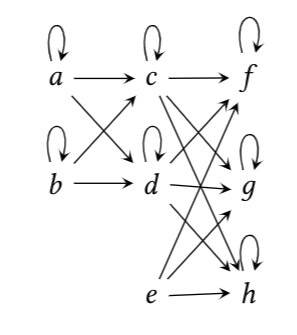}
%  \caption{Two consecutive fully connected layers. The first one without bias and the second one with bias.}
%\end{subfigure}
\end{center}

\caption{(a) Combinatorial architecture of a fully connected layer. This architecture has no restrictions on the weight nor the activation architectures. (b) Two consecutive fully connected layers. The first without bias and the second with bias. Note that there is no loop (activation function) on the bias vertex.} \label{fig:FC_layers}
\end{figure}

%\subsection{Fully connected layers}
Let $\V^j$ be the set of vertices on the $j$-th hidden layer of $Q$. A \textbf{fully connected layer} is a hidden layer $\V^j$ where all vertices on the previous layer are connected to all vertices in $\V^j$. A \textbf{fully connected layer with bias} is a hidden layer $\V^j$ that puts constraints on the previous layer $\V^{j-1}$ such that the non-bias vertices of $\V^{j-1}$ are fully connected with the non-bias vertices of layer $\V^j$. % Secondly, there is a bias vertex in $\V^{j-1}$ connected to every vertex of $\V^{j}$. 
A fully connected layer has no constraints on its weight and activation architecture but impose that the bias vertex has no activation function and not connected with the vertex of the previous layer. The reader can find an illustration of this in Fig.~\ref{fig:FC_layers}.

%\subsubsection{Convolutional layers}
 A \textbf{convolutional layer} is a hidden layer $\V^j$ whose vertices are separated in \text{channels} (or feature maps).  The weights are typically organized in filters $(F_n)_{n=1}^m$, and each $F_n$ is a tensor {\color{black}made of} channels. {\color{black}By ``channels'', we mean that the shape of, for example, a 2D convolution is given by $w \times h \times c$, where $w$ is the width, $h$ is the height and $c$ is the number of channels on the previous layer.  %, and this is what we mean by ``partition''. 
 A ``filter'' is given by the weights and edges on a conv layer whose target lies in the same channel.}
 
\begin{figure} 
%\centering
%\begin{subfigure}{.5\textwidth}
%  \centering
(a)  \includegraphics[width=.19\linewidth]{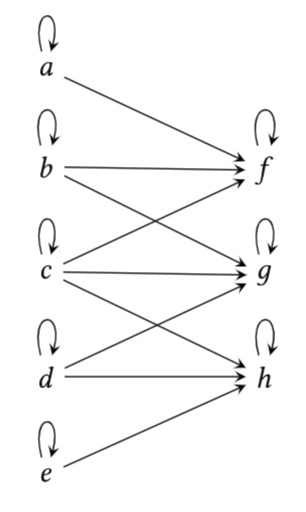}
%  \caption{The combinatorial architecture of a convolutional layer.}
%\end{subfigure}%
%\begin{subfigure}{.5\textwidth}
%  \centering
(b)  \includegraphics[width=.19\linewidth]{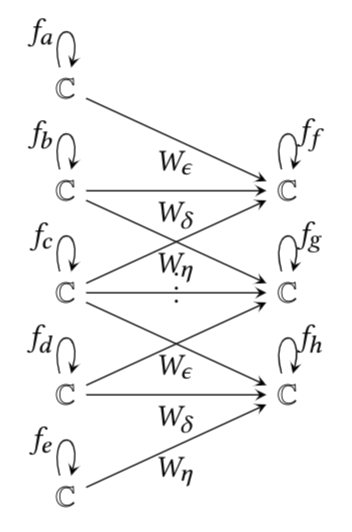}
%  \caption{The weight architecture of a convolutional layer.}
%\end{subfigure}
(c)  \includegraphics[width=.19\linewidth]{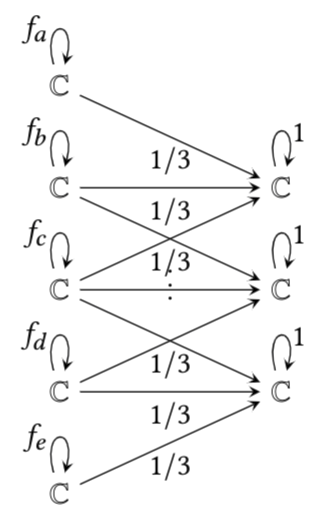}
(d)  \includegraphics[width=.23\linewidth]{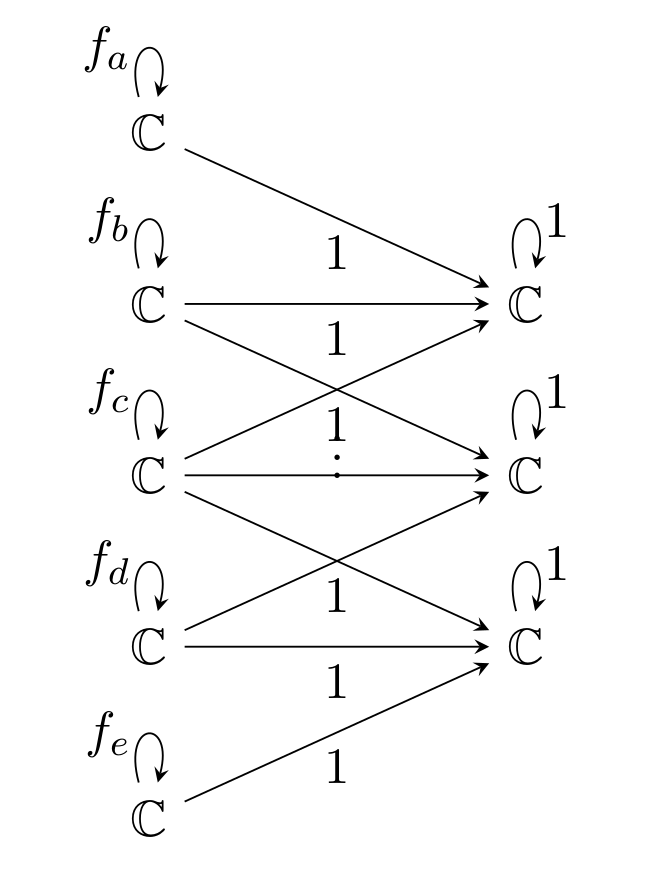}
\caption{(a) Combinatorial architecture of a convolutional and a pooling layer. (b) Weight and activation architecture of a convolutional layer. (c) Weight and activation architecture of an average pooling layer. (d) Weight and activation architecture of a max-pooling layer.} \label{fig:conv_pool_layers}
\end{figure}

 {\color{black}As opposed to fully-connected layers, convolutional layers have constraints.  One of which is that convolutional layers should be partitioned into  channels of the same cardinality. Each filter $F_n$ produces a channel on the layer $\V^j$ by a convolution of $\V^{j-1}$ with the filter $F_n$.  Also, a convolution operation has a stride and may use padding.
 
 A convolutional layer also has constraints on its combinatorial and weight architecture.}  First, each $\V^j$ is connected to a sub-set of vertices in the previous layer ``in front" of which it is located.  The combinatorial architecture of a conv layer for one feature map is illustrated in Fig.~\ref{fig:conv_pool_layers}(a). Second, the weight architecture requires that the weights on the filters repeat in every sliding of the convolutional window.  In other words, the weights of the edges on a conv layer must be shared across all filters as in Fig.~\ref{fig:conv_pool_layers}(b).
A \textbf{conv layer with bias} is a hidden layer $\V^j$ partitioned into channels, where each channel is obtained by convolution of $\V^{j-1}$ with each filter $F_n$, $n=1,...,m$, plus one bias vertex in layer $\V^{j-1}$ that is connected to every vertex on every channel of $\V^j$. The weights of the edges starting on the bias vertex should repeat within the same channel.  Again, bias vertices do not have an activation function and are not connected to neurons of the previous layer.%Again we can see that this type of layer concerns both weight architecture and combinatorial architecture.

%\subsubsection{Pooling layers}

The combinatorial architecture of a \textbf{pooling layer} is the same as that of a conv layer, see Fig.~\ref{fig:conv_pool_layers}(a).  However, since the purpose of that operation is usually to reduce the size of the previous layer, it contains non-trainable parameters.  Thus, pooling layers have a different weight architecture than the conv layers.
Average pooling fixes the weights in a layer to $1/n$ where $n$ is the size of the feature map, while max-pooling {\color{black} fixes the weights in a layer to $1$ and outputs the maximum over each window in the previous layer}.  Also, the activation function of an average and max-pooling layer is the identity function. This can be appreciated in Fig.~\ref{fig:conv_pool_layers}(c) and (d). 
%\[
%\begin{array}{ccc}
% \begin{tikzpicture}
%         	\matrix (m) [matrix of math nodes,row sep=2em,column sep=2em]
%          	{
%           	  	 a & & \\
%           	  	 b & & \\
%           	  	 \vdots & & g \\
%           	  	 f & & \\
%           	};
%          	\path[-stealth]
%      	 (m-1-1) edge node [right] {} (m-3-3)
%   	     (m-2-1) edge node [left] {} (m-3-3)
%       	 (m-4-1) edge node [right] {} (m-3-3);%
%	\end{tikzpicture}    & \begin{tikzpicture}
%         	\matrix (m) [matrix of math nodes,row sep=2em,column sep=2em]
%          	{
%           	  	 \Cmplx & & \\
%           	  	 \Cmplx & & \\
%           	  	 \vdots & & \Cmplx \\
%           	  	 \Cmplx & & \\
%           	};
%          	\path[-stealth]
%      	 (m-1-1) edge node [right] {$1/n$} (m-3-3)
%   	     (m-2-1) edge node [left] {$1/n$} (m-3-3)
%       	 (m-4-1) edge node [right] {$1/n$} (m-3-3);
%	\end{tikzpicture} & \text{( resp. } \begin{tikzpicture}
%         	\matrix (m) [matrix of math nodes,row sep=2em,column sep=2em]
%          	{
%           	  	 \Cmplx & & \\
%           	  	 \Cmplx & & \\
%           	  	 \vdots & & \Cmplx \\
%           	  	 \Cmplx & & \\
%           	};
%          	\path[-stealth]
%      	 (m-1-1) edge node [below] {$0$} (m-3-3)
%   	     (m-2-1) edge node [below] {$1$} (m-3-3)
%       	 (m-4-1) edge node [below] {$0$} (m-3-3);
%	\end{tikzpicture} \\
%\end{array}
%\]
%where the $1$ is in the place of the maximum).

\begin{figure} 
%\centering
%\begin{subfigure}{.5\textwidth}
%  \centering
(a)  \includegraphics[width=.25\linewidth]{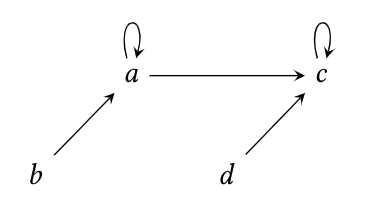}
%  \caption{The combinatorial architecture of a convolutional layer.}
%\end{subfigure}%
%\begin{subfigure}{.5\textwidth}
%  \centering
(b)  \includegraphics[width=.25\linewidth]{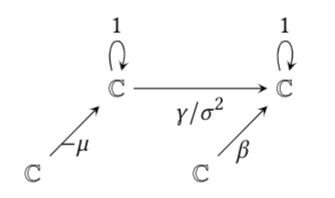}
%  \caption{The weight architecture of a convolutional layer.}
%\end{subfigure}
\caption{(a) Combinatorial architecture of a batch normalization layer. (b) Weight and activation architecture of a batch normalization layer. {\color{black} Observe that vertices $b$ and $d$ are bias vertices, and therefore the layer computes $x \mapsto x-\mu \mapsto (x-\mu)(\gamma/\sigma^2)+\beta$, which is the definition of the batch norm operation.}} \label{fig:bn_layers}
\end{figure}

\begin{Remark} \label{rmk:pooling}
    Max-pooling layers are compatible with our constructions, but they force us to consider {\color{black} another operation in the neuron, as was noted in Definition~\ref{def:actfun}}. % neural networks $(W,f)$ where the weights on the max-pooling layer are not fixed until the network is fed with an input. This complicates the algebraic methods needed to model it as we need to consider sets of neural networks (one for each possibility of weights in the max-pooling layer) instead of just one.
    
    It is known that max-pooling layers give small amount of translation invariance at each level since the precise location of the most active feature detector is thrown away, and this produces doubts about the use of max-pooling layers~\citep[see][]{Hinton14vid,Sabour17}. An alternative to this is the use of attention-based pooling~\citep{Kosiorek19}, which is a global-average pooling.  Our interpretation provides a framework that supports why these doubts about the use of max-pooling layers exist: they break the algebraic structure {\color{black}on the computations} of a neural network.  However, average pooling layers, and therefore global-average pooling layers, are perfectly consistent with respect to our results since they are given by fixed weights for any input vector {\color{black} while not requiring specification of another operation}.
    %This puts our interpretation in sync with current intuition on pooling layers for the case of max-pooling~\citep[see][]{Hinton14vid}. Nevertheless, our language completely supports the use of average and global-average pooling layers~\citep{Kosiorek19}.
\end{Remark}
%\subsubsection{Batch normalization layers}
\textbf{Batch normalization layers}~\citep{Ioffe15} require specifications on the three types of architecture. Their combinatorial architecture is given by two identical consecutive hidden layers where each neuron on the first is connected to only one neuron on the second, and there is one bias vertex in each layer. The weight architecture is given by the batch norm operation, which is $x \mapsto \dfrac{x-\mu}{\sigma^2} \gamma + \beta$ where $\mu$ is the mean of a batch and $\sigma^2$ its variance, and $\gamma$ and $\beta$ are learnable parameters.  The activation architecture is given by two identity activations. This can be seen in Fig.~\ref{fig:bn_layers}. %This neuron-wise operation $\Cmplx \to \Cmplx$ is represented by the diagram:

%\[
%	\begin{tikzpicture}
%         	\matrix (m) [matrix of math nodes,row sep=2em,column sep=2em]
%          	{
%           	  	  & a & & c \\
%           	  	 b & & d &  \\
%           	};
%          	\path[-stealth]
%   	     (m-1-2) edge node [below] {} (m-1-4)
%       	 (m-2-1) edge node [below] {} (m-1-2)
%       	 (m-2-3) edge node [below] {} (m-1-4)
%       	 (m-1-2) edge [loop above] node {} (m-1-2)
%       	 (m-1-4) edge [loop above] node {} (m-1-4);
%	\end{tikzpicture}
%\]
%\[
%	\begin{tikzpicture}
%         	\matrix (m) [matrix of math nodes,row sep=2em,column sep=2em]
%          	{
%           	  	  & \Cmplx & & \Cmplx \\
%           	  	 \Cmplx & & \Cmplx &  \\
%           	};
%          	\path[-stealth]
%   	     (m-1-2) edge node [below] {$\gamma/\sigma^2$} (m-1-4)
%       	 (m-2-1) edge node [below] {$-\mu$} (m-1-2)
%       	 (m-2-3) edge node [below] {$\beta$} (m-1-4)
%       	 (m-1-2) edge [loop above] node {$1$} (m-1-2)
%       	 (m-1-4) edge [loop above] node {$1$} (m-1-4);
%	\end{tikzpicture}
%\]
%where the bottom vertices are bias vertices.

\begin{Remark} %\label{rmk:bn}
    The weights $\mu$ and $\sigma$ {\color{black}are not} determined until the network is fed with a batch of data.  %But that is true only at training time.  
    {\color{black} However, at test time}, $\mu$ and $\sigma$ are set to the overall mean and variance computed across the training data set and thus become normal weights. {\color{black} This does not means that the architecture of the network depends on the input vector, but that the way these particular weights are chosen is by obtaining mean and variance from the data.}
\end{Remark}

\begin{figure} 
%\centering
%\begin{subfigure}{.5\textwidth}
\begin{center}
(a)  \includegraphics[width=.45\linewidth]{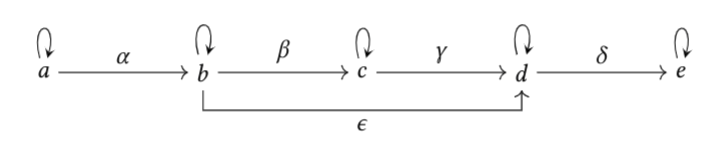} 
%  \caption{The combinatorial architecture of a convolutional layer.}
%\end{subfigure}%
%\begin{subfigure}{.5\textwidth}
%  \centering
  (b) \includegraphics[width=.45\linewidth]{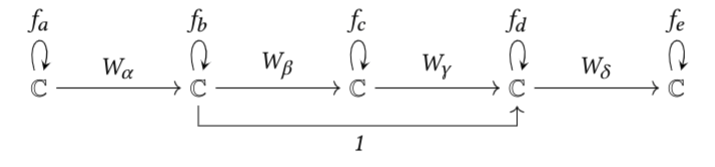}
\end{center}
%  \caption{The weight architecture of a convolutional layer.}
%\end{subfigure}
\caption{(a) Combinatorial architecture of a residual connection. (b) Weight architecture of a residual connection.} \label{fig:residual_layers}
\end{figure}

%\subsubsection{Residual connections}
The combinatorial architecture of a \textbf{residual connection}~\citep{He16} requires the existence of edges in $Q$ that jump over one or more layers. Their weight architecture forces the weights chosen for those edges to be always equal to $1$. We refer to Fig.~\ref{fig:residual_layers} for an illustration of the architecture of a residual connection.

%For example, the next thin representation of a delooped quiver $Q^\circ$ has a residual connection:
%\[
%\begin{tikzpicture}[auto]
%   \node (A) at (1,9) {$\Cmplx$};
%   \node (B) at (3,9) {$\Cmplx$};
%   \node (C) at (5,9) {$\Cmplx$};
%   \node (D) at (7,9) {$\Cmplx$};
%   \node (E) at (9,9) {$\Cmplx.$};
%   \draw[->] (A) to  node {$W_\alpha$} (B);
%   \draw[->] (B) to  node {$W_\beta$} (C);
%   \draw[->] (C) to  node {$W_\gamma$} (D);
%   \draw[->] (D) to  node {$W_\delta$} (E);
%   \draw[->] (B) to [ncbar=-1.5em] node[pos=0.5,below] {1} (D); 
%\end{tikzpicture}
%\]

\subsection{Architecture preserved by isomorphisms} \label{sec:inv-isom}
Two isomorphic neural networks can have different weight architectures. Let us illustrate this with a residual connection. Let $Q$ be the following network quiver
\[
    \begin{tikzpicture}[auto]
       \node (A) at (1,9) {$a$};
       \node (B) at (3,9) {$b$};
       \node (C) at (5,9) {$c$};
       \node (D) at (7,9) {$d$};
       \node (E) at (9,9) {$e$};
       \path[-stealth] 
       %(A) edge [loop above] node {} (A)
       (B) edge [loop above] node {} (B)
       (C) edge [loop above] node {} (C)
       (D) edge [loop above] node {} (D);
       %(E) edge [loop above] node {} (E);
       \draw[->] (A) to  node {$\alpha$} (B);
       \draw[->] (B) to  node {$\beta$} (C);
       \draw[->] (C) to  node {$\gamma$} (D);
       \draw[->] (D) to  node {$\delta$} (E);
       \draw[->] (B) to [ncbar=-1.5em] node[pos=0.5,below] {$\epsilon$} (D); 
    \end{tikzpicture}
\]
and the neural network $(W,f)$ over $Q$ given by
\[
    \begin{tikzpicture}[auto]
       \node (A) at (1,9) {$\Cmplx$};
       \node (B) at (3,9) {$\Cmplx$};
       \node (C) at (5,9) {$\Cmplx$};
       \node (D) at (7,9) {$\Cmplx$};
       \node (E) at (9,9) {$\Cmplx.$};
       \path[-stealth] 
       %(A) edge [loop above] node {$f_a$} (A)
       (B) edge [loop above] node {$f_b$} (B)
       (C) edge [loop above] node {$f_c$} (C)
       (D) edge [loop above] node {$f_d$} (D);
       %(E) edge [loop above] node {$f_e$} (E);
       \draw[->] (A) to  node {$W_\alpha$} (B);
       \draw[->] (B) to  node {$W_\beta$} (C);
       \draw[->] (C) to  node {$W_\gamma$} (D);
       \draw[->] (D) to  node {$W_\delta$} (E);
       \draw[->] (B) to [ncbar=-1.5em] node[pos=0.5,below] {1} (D); 
    \end{tikzpicture}
\]

Let $\tau_b \not= \tau_d$ be non-zero numbers, we define a change of basis of the neural network $(W,f)$ by $\tau=(1,\tau_b,1,\tau_d,1)$. After applying the action of the change of basis $\tau \cdot (W,f)$ we obtain an isomorphic neural network given by
    \[
    \begin{tikzpicture}[auto]
       \node (A) at (1,9) {$\Cmplx$};
       \node (B) at (3,9) {$\Cmplx$};
       \node (C) at (5,9) {$\Cmplx$};
       \node (D) at (7,9) {$\Cmplx$};
       \node (E) at (9,9) {$\Cmplx.$};
       \path[-stealth] 
       %(A) edge [loop above] node {$f_a$} (A)
       (B) edge [loop above] node {$\tau_b \cdot f_b$} (B)
       (C) edge [loop above] node {$f_c$} (C)
       (D) edge [loop above] node {$\tau_d \cdot f_d$} (D);
       %(E) edge [loop above] node {$f_e$} (E);
       \draw[->] (A) to  node {$W_\alpha \tau_b$} (B);
       \draw[->] (B) to  node {$W_\beta / \tau_b$} (C);
       \draw[->] (C) to  node {$W_\gamma \tau_d$} (D);
       \draw[->] (D) to  node {$W_\delta / \tau_d$} (E);
       \draw[->] (B) to [ncbar=-1.5em] node[pos=0.5,below] {$\tau_b/\tau_d$} (D); 
    \end{tikzpicture}
    \]
The neural networks $(W,f)$ and $\tau \cdot (W,f)$ are isomorphic and therefore they have the same network function by Theorem~\ref{thm:netfunc}. However, the neural network $(W,f)$ has a residual connection, while $\tau \cdot (W,f)$ does not since the weight on the skip connection is not equal to 1. Nevertheless, if we take $\tau_b=\tau_d$, then the change of basis $\tau' = (1,\tau_b,1,\tau_b,1)$ will produce an isomorphic neural network with a residual connection, and therefore both neural networks $(W,f)$ and $\tau' \cdot (W,f)$ will have the same weight architecture.

The same phenomenon as for residual connections happens for convolutions, where one has to choose a specific kind of isomorphism to preserve the weight architecture, as shown in Fig.~\ref{fig:tel_conv}. Isomorphisms of neural networks preserve the combinatorial architecture but not necessarily the weight architecture nor the activation architecture. 

\subsection{Consequences}% and possible outcomes}

As for the previous section, expressing neural network layers through the basic definitions of quiver representation theory has some consequences. Let us mention two.

\paragraph{Consequence 1}
The first consequence derives from the isomorphism of residual layers.
%\begin{Remark}
    It is claimed by~\citet{Meng18} that there is no positive scale invariance across residual blocks. However, we can see that the quiver representation language allows us to prove that in fact there is positive scale invariance across residual blocks for ReLU networks. {\color{black} Therefore, isomorphisms allow to understand that there are far more symmetries on neural networks than was previously known, as noted in Section~\ref{sec:inv-isom}, which can be written as follows:
%\end{Remark}
\begin{Corollary}
    There is invariance across residual blocks under isomorphisms of neural networks.
\end{Corollary}
}

\begin{figure} 
  \includegraphics[width=.5\linewidth]{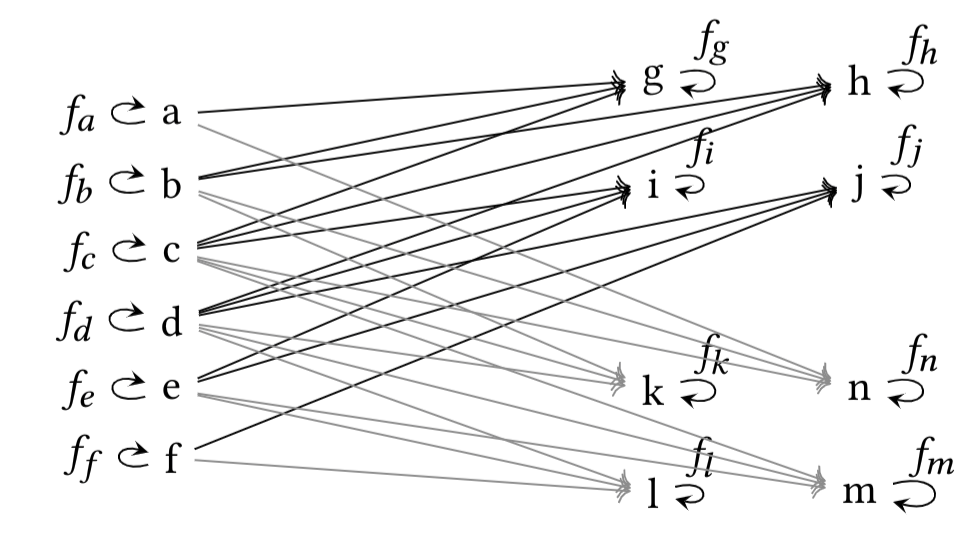} 
\caption{An illustration of a convolutional layer. The black arrows with target g, h, i and j correspond to the first channel, and the gray arrows with target k, l, m and n correspond to the second channel. A change of basis $\tau \in \Gt$ that preserves the weight architecture of this convolutional layer, has to be of the form $\tau=(\tau_i)_{i=a}^{m}$ where $\tau_g=\tau_h=\tau_i=\tau_{j}$ and $\tau_{k}=\tau_{l}=\tau_{m}=\tau_{n}$. {\color{black}Note that a convolution is given by filters which share weights, so if the previous condition is not satisfied, after applying the change of basis one will obtain weights that are not shared, so the resulting weights will not fit the definition of a convolution.}%This is what is done in \citep{Meng18}, for the particular case of ReLU networks and positive change of basis. 
} \label{fig:tel_conv}
\end{figure}

\paragraph{Consequence 2}
The second consequence is related to the existence of isomorphisms that preserve the weight architecture and not the activation architecture.  As in Fig.~\ref{fig:tel_conv}, a change of basis $\tau \in \Gt$ that preserves the weight architecture of this convolutional layer, has to be of the form $\tau=(\tau_i)_{i=a}^{m}$ where $\tau_g=\tau_h=\tau_i=\tau_{j}$ and $\tau_{k}=\tau_{l}=\tau_{m}=\tau_{n}$. This is what~\citet{Meng18} do for the particular case of ReLU networks and positive change of basis {\color{black}(they consider the action of the group $\prod_{v \in \widetilde{Q}} \mathbb{R}_{>0}$ on neural networks)}. {\color{black} Note that if the change of basis is not chosen in this way, the isomorphism will produce a layer with different weights in each convolutional filter, and therefore the resulting operation will not be a convolution with respect to the same filter.} While positive scale invariance of ReLU networks is a special kind of invariance under isomorphisms of neural networks that preserve both the weight and the activation architecture, we may generalize this notion {\color{black}by allowing isomorphisms to change the activation architecture while preserving the weight architecture}.
\begin{Definition}
    Let $(W,f)$ be a neural network and let $\tau \in \Gt$ be an element of the group of change of basis of neural networks such that the isomorphic neural network $\tau \cdot (W,f)$ has the same weight architecture as $(W,f)$. The \textbf{teleportation} of the neural network $(W,f)$ with respect to $\tau$ is the neural network $\tau \cdot (W,f)$.  
\end{Definition}
%Please refer to the Appendix~\ref{Appendix2} for a simple example illustrating the teleportation  $\tau \cdot (W,f)$ of a multilayer perceptron.
%\begin{Remark}
    {\color{black}Since teleportation preserves the weight architecture, it follows that the teleportation of a conv layer is a conv layer, the teleportation of a pooling layer is a pooling layer, the teleportation of a batch norm layer is a batch norm layer, and the teleportation of a residual block is a residual block.} Teleportation produces a neural network with the same combinatorial architecture, weight architecture and network function while it may change the activation architecture. For example, consider a neural network with ReLU activations and real change of basis. Since ReLU is positive scale invariant, any positive change of basis will leave ReLU invariant. On the other hand, for a negative change of basis the activation function changes to $min(0,x)$ and therefore the weight optimization landscape also changes. This implies that teleportation may change the optimization problem {\color{black}by changing the activation functions,} while preserving the network function, and the network gets ``\textit{teleported}'' to either other place in the same loss landscape {\color{black}(if the activation functions are not changed)} or to a completely different loss landscape {\color{black}(if activation functions are changed)}.
%\end{Remark}
%\begin{Proposition}
%    Let $(W,f)$ be a neural network over $Q$ and let $H$ be the biggest sub-group of the change of basis group $\Gt$ given by the elements $\tau \in \Gt$ such that $\tau \cdot f=f$ and $\tau$ defines a teleportation. If $f$ is ReLU or $f(x)=min(x,0)$, then
%    \[
%        H = \displaystyle\prod_{i=1}^{\# \\V} \R_{>0}.
%    \]
%\end{Proposition}
% {\bf Proof}.
%    Indeed, the group whose action commutes with ReLU and $f(x)=min(x,0)$ is the multiplicative group of positive real numbers.
% 

\section{Data Representations}
\label{sec:datarep}

In machine learning, a data sample is usually represented by a vector, a matrix or a tensor containing a series of observed variables.  However, one may view data from a different perspective, namely the neuron outputs obtained after a forward pass, also known as ``feature maps" for conv nets~\citep{Goodfellow-et-al-2016}.  This has been done in the past to visualize what neurons have learned~\citep{Hinton07,Bengio13,Yosinski15}.  

In this section, we propose %how a neural network computes a forward pass for each data sample in order to obtain 
a mathematical description of the data in terms of the architecture of the neural network, i.e., the neuron values obtained after a forward pass.  We shall prove that doing so allows to represent data by a quiver representation.  Our approach is different from representation learning~\citep[page~4]{Goodfellow-et-al-2016} because we do not focus on how the representations are learned but rather on how the representations of the data are encoded by the forward pass of the neural network.

\begin{Definition}
    %Let $Q$ be a network quiver. 
    A \textbf{labeled data set} % for $Q$ 
    is given by a finite set $D = \{ (x_i,t_i) \}_{i=1}^n$ of pairs such that $x_i \in \Cmplx^d$ is a data vector (could also be a matrix or a tensor) and $t_i$ is a target. We can have $t_i \in \Cmplx^k$ for a regression and $t_i \in \{C_0,C_1,...,C_k\}$ for a classification.
\end{Definition}
%\begin{Remark}
%    In applications, a dataset $\dataset$ is given by pairs such that $x_i \in \R^d$ and $y_i \in \R^K$. One also partitions the dataset $\dataset$ into two sets, the training set and the test set. The points $y_i \in \R^K$ are only used either to tell something to the optimization algorithm or to measure the accuracy or loss of a neural network. Since we are putting the optimization aside, we will not be working with the second coordinates of data samples. However, we will strongly address the optimization-quiver theoretic interaction in future work.
%\end{Remark}
 
Let $(W,f)$ be a neural network over a network quiver $Q$ and a sample $(x,t)$ of a data set $D$. When the network processes the input $x$, the vector $x$ percolates through the edges and the vertices from the input to the output of the network.  As mentioned before, this results in neuron values (or feature maps) that one can visualize~\citep{Yosinski15}.  On its own, the neuron values are not a quiver representation per se.  However, one can combine these neuron values with their pre-activations and the network weights to obtain \textbf{a thin quiver representation}.  Since that representation derives from the forward pass of $x$, it is specific to it. {\color{black}We will evaluate the activation functions in each neuron and then construct with them a quiver representation for a given input. We stress out that this process is not ignoring the very important non-linearity of the activation functions, so no information of the forward pass is lost in this interpretation.}

%Our data representation is based on these neuron vales that we express as a thin quiver representation $\Wxf$ of the delooped quiver $Q^\circ$.% that will keep track of every computation of the network while keeping the output of the network invariant.

%If we were to keep track of every computation of the network, can we put all of those computations into a mathematical object so that the output remains invariant? It turns out we can, by using quiver representations.

%that can harness this pouring of the vector $x$ into the network together with the activation outputs of every neuron

%The idea comes from what happens on bias vertices when we process data, namely that the bias vertices are provided with an input of $1$. We want to put the input vector $x$ into the network, so that we can consider a canonical input of only ones for the quiver representation that will encode the computations of the network. For this, we construct a thin representation $\Wxf$ of the delooped quiver $Q^\circ$.
\begin{Remark}
    Every thin quiver representation $V$ of the delooped quiver $Q^\circ$ defines a neural network over the network quiver $Q$ with identity activations, that we denote $(V,1)$.  We do not claim that taking identity activation functions for a neural network will result in something good in usual deep learning practices.  This is only a theoretical trick to manipulate the underlying algebraic objects we have constructed. {\color{black}As such, we will identify thin quiver representations $V$ with neural networks with identity activation functions $(V,1)$}.
\end{Remark}

%Now let's consider the score of that network $\Psi (W,f)(x)$ after a forward pass.  
Our data representation for $x$ is a thin representation that we call $\Wxf$ with identity activations whose function when fed with an input vector of ones $1^d:=(1,...,1) \in \Cmplx^d$ satisfies
\begin{eqnarray}
\Psi (\Wxf,1)(1^d) = \Psi (W,f)(x),
\end{eqnarray}
where $\Psi (W,f)(x)$ is the score of the network $(W,f)$ after a forward pass of $x$.

\begin{figure} 
\centering
%\begin{subfigure}{.5\textwidth}
%  \centering
  (a)\includegraphics[width=.44\linewidth]{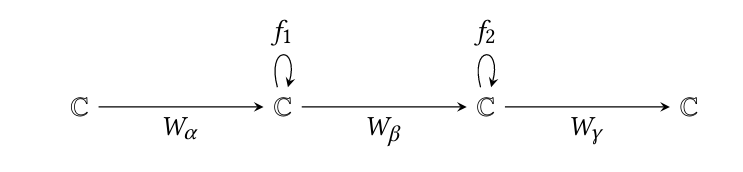} 
%  \caption{A quiver.}
%\end{subfigure}%
%\begin{subfigure}{.5\textwidth}
%  \centering
(b)  \includegraphics[width=.4\linewidth]{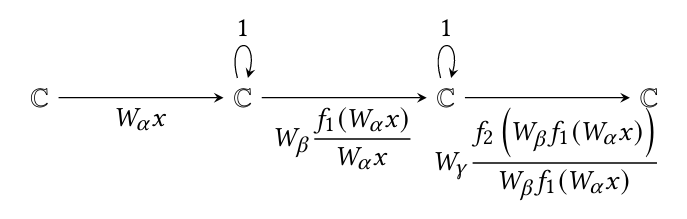}  \\
(c)  \includegraphics[width=.4\linewidth]{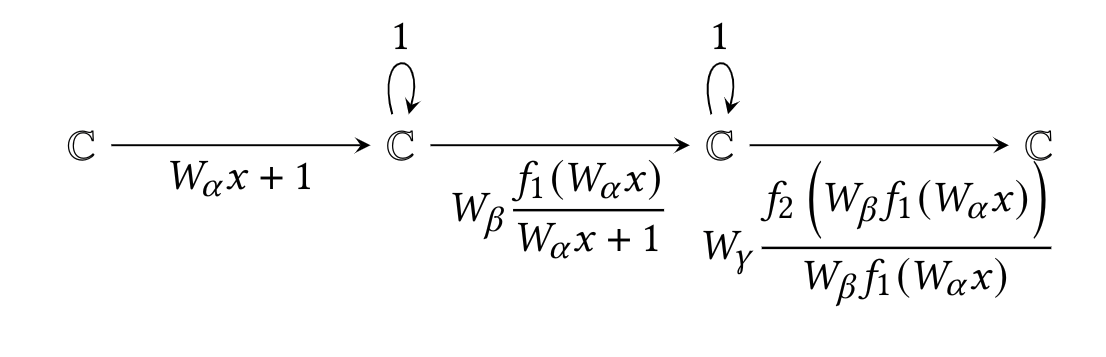} 
\caption{(a) A neural network (W,f). (b) The induced thin quiver representation $\Wxf$ considered as a neural network $(\Wxf,1)$ and obtained after feed-forwarding $x$ through $(W,f)$. It can be seen that feed-forwarding a unit vector $1$ through $\Wxf$ (i.e., $\Psi(\Wxf,1)(1)$) gives the same output than feed-forwarding $x$ through $(W,f)$ : $\Psi(\Wxf,1)(1) = \Psi(W,f)(x)$. We refer to Theorem~\ref{thm:data} for the general case. (c) In the case $W_\alpha x =0$, we can add 1 to the corresponding pre-activation in $\Wxf$ to prevent from a division by zero, while on the next layer we consider $W_\alpha x + 1$ as the pre-activation.} \label{fig:wxf}
\end{figure}

Recovering $\Wxf$ given the forward pass of $x$ through $(W,f)$ is illustrated in Fig.~\ref{fig:wxf} (a) and (b).
Let's keep track of the computations of the network in the thin quiver representation $\Wxf$ and remember that at the end, we want the output of the neural network $(\Wxf,1)$ when fed with the input vector $1^d \in \Cmplx^d$, to be equal to $\Psi(W,f)(x)$.

If $\epsilon \in \mathcal{E}$ is an oriented edge such that $s(\epsilon) \in \V$ is a bias vertex, then the computations of the weight corresponding to $\epsilon$ get encoded as $\left( \Wxf \right)_\epsilon= W_\epsilon$. If on the other hand % $\epsilon \in \mathcal{E}$ is an oriented edge such that 
$s(\epsilon) \in \V$ is an input vertex, then the computations of the weights on the first layer get encoded as $\left( \Wxf \right)_\epsilon= W_\epsilon x_{s(\epsilon)}$, see Fig.~\ref{fig:wxf}(b).

On the second and subsequent layers of the network $(W,f)$ we encounter activation functions.  Also, the weight corresponding to an oriented edge $\epsilon$ in $\Wxf$ will have to cancel the unnecessary computations coming from the previous layer.  That is, $\left( \Wxf \right)_\epsilon$ has to be equal to $W_\epsilon$ times the activation output of the vertex $s(\epsilon)$ divided by the pre-activation of $s(\epsilon)$. %An example of how this works can be found in Fig.~\ref{fig:wxf})(a),(b).  
Overall, $\Wxf$ is defined as

\begin{equation} \label{Wxf}
    \left( \Wxf \right)_\epsilon = \left\{    \begin{array}{ll}
        W_\epsilon x_{s(\epsilon)} & \text{ if } s(\epsilon) \text{ is an input vertex,}  \\
        W_\epsilon & \text{ if } s(\epsilon) \text{ is a bias vertex,} \\
        W_\epsilon \dfrac{\act(W,f)_{s(\epsilon)}(x) }{  \displaystyle\sum_{\beta \in \zeta_{s(\epsilon)}}  W_\beta \cdot \act(W,f)_{s(\beta)}(x)} & \text{ if } s(\epsilon) \text{ is a hidden vertex, } \\
    \end{array} \right.
\end{equation}
where $\zeta_{s(\epsilon)}$ is the set of oriented edges of $Q$ with target $s(\epsilon)$.  In the case where the activation function is ReLU, for $\epsilon$ an oriented edge such that $s(\epsilon)$ is a hidden vertex, either $\left( \Wxf \right)_{\epsilon} = 0 $ or $\left( \Wxf \right)_{\epsilon} = W_\epsilon $.

\begin{Remark}
    Observe that the denominator $\displaystyle\sum_{\beta \in \zeta_{s(\epsilon)}}  W_\beta \cdot \act(W,f)_{s(\beta)}(x)$ is the pre-activation of vertex $s(\epsilon)$ and can be equal to zero. However, the set where this happens is of measure zero.  And even in the case that it turns out to be exactly zero, one can add a number $\eta\not=0$ (for example $\eta=1$) to make it non-zero and then consider $\eta$ as the pre-activation of that corresponding neuron, see Fig.~\ref{fig:wxf}(c).  So we will assume, without loss of generality, that pre-activations of neurons are always non-zero.
\end{Remark}
The quiver representation $\Wxf$ of the delooped quiver $Q^\circ$ accounts for the combinatorics of the history of all the computations that the neural network $(W,f)$ performs on a forward pass given the input $x$. The main property of the quiver representation $\Wxf$ is given by the following result. A small example of the computation of $\Wxf$ and a view into how the next Theorem works, can be found in Appendix~\ref{appendix2}.
\begin{Theorem} \label{thm:data}
    Let $(W,f)$ be a neural network over $Q$, let $(x,t)$ be a data sample for $(W,f)$ and consider the induced thin quiver representation $\Wxf$ of $Q^\circ$. The network function of the neural network $(\Wxf,1)$ satisfies
    \[
        \Psi(\Wxf,1)(1^d)=\Psi(W,f)(x).
    \]
\end{Theorem}
 {\bf Proof}.
    Obviously, both neural networks have different input vectors, that is, $1^d$ for $(\Wxf,1)$ and $x$ for $(W,f)$.  If $v\in \V$ is a source vertex, by definition $\act(\Wxf,1)_v(1^d) = 1$.  We will show that in the other layers, the activation output of a vertex in $(\Wxf,1)$ is equal to the pre-activation of $(W,f)$ in that same vertex. Assume that $v\in \V$ is in the first hidden layer, let $\zeta_v^{bias}$ be the set of oriented edges of $Q$ with target $v$ and source vertex a bias vertex, and let $\zeta_v^{input}$ be the set of oriented edges of $Q$ with target $v$ and source vertex an input vertex. Then, for every $\epsilon \in \zeta_{v}$ where $\zeta_v=\zeta_v^{bias}\cup \zeta_v^{input}$, we have that $\act(\Wxf,1)_{s(\epsilon)}(1^d) = 1$, and therefore
    \[
        \begin{array}{lcl}
            \act(\Wxf,1)_v(1^d) & = & \displaystyle\sum_{\epsilon \in \zeta_v} \left( \Wxf \right)_\epsilon \act(\Wxf,1)_{s(\epsilon)} (1^d) \\
             & = & \displaystyle\sum_{\epsilon \in \zeta_v} \left( \Wxf \right)_\epsilon \\
             & = & \displaystyle\sum_{\epsilon \in \zeta_v^{bias} } \left( \Wxf \right)_\epsilon  + \displaystyle\sum_{\epsilon \in \zeta_v^{input} } \left( \Wxf \right)_\epsilon  \\
             & = & \displaystyle\sum_{\epsilon \in \zeta_v^{bias} } W_\epsilon  + \displaystyle\sum_{\epsilon \in \zeta_v^{input} } W_\epsilon x_{s(\epsilon)}, \\
        \end{array}
    \]
    which is the pre-activation of vertex $v$ in $(W,f)$, i.e., $f_v\left( \act(\Wxf,1)_v(1^d) \right) = \act(W,f)_v(x)$. If $v\in \V$ is in the second hidden layer then
    \[
        \begin{array}{lcl}
            \act(\Wxf,1)_v(1^d) & = & \displaystyle\sum_{\epsilon \in \zeta_v} \left( \Wxf \right)_\epsilon \act(\Wxf,1)_{s(\epsilon)} (1^d)  \\
             & = & \displaystyle\sum_{\epsilon \in \zeta_v} W_\epsilon \dfrac{\act(W,f)_{s(\epsilon)}(x) }{  \displaystyle\sum_{\beta \in \zeta_{s(\epsilon)}}  W_\beta \cdot \act(W,f)_{s(\beta)}(x)} \act(\Wxf,1)_{s(\epsilon)} (1^d) \\
        \end{array}
    \]
    since $\displaystyle\sum_{\beta \in \zeta_{s(\epsilon)}}  W_\beta \cdot \act(W,f)_{s(\beta)}(x)$ is the pre-activation of vertex $s(\epsilon)$ in $(W,f)$, by the above formula we get that $\displaystyle\sum_{\beta \in \zeta_{s(\epsilon)}}  W_\beta \cdot \act(W,f)_{s(\beta)}(x) = \act(\Wxf,1)_{s(\epsilon)} (1^d)$, and then
    \[
        \begin{array}{lcl}
           \act(\Wxf,1)_v(1^d) & = & \displaystyle\sum_{\epsilon \in \zeta_v} W_\epsilon \act(W,f)_{s(\epsilon)}(x), \\
        \end{array}
    \]
    which is the pre-activation of vertex $v$ in $(W,f)$ when fed with the input vector $x$. That is, $f_v\left( \act(\Wxf,1)_v(1^d) \right) = \act(W,f)_v(x)$. An induction argument gives the desired result since the output layer has no activation function, and the coordinates of $\Psi(\Wxf,1)(1^d)$ and $\Psi(W,f)(x)$ are the values of the output vertices.\hfill\BlackBox

    %On the other side, if $\Psi(W,f)(x) = \Psi(W,f)(x')$ then by the last Theorem $\Psi(\Wxf,1)(1^d) = \Psi(\texttt{W}_{x'}^f,1)(1^d)$. Since the vector $1^d$ is the sum of the canonical basis vectors of $\Cmplx^d$ we get that the matrices defining the linear maps $\Psi(\Wxf,1)$ and $\Psi(\texttt{W}_{x'}^f,1)$ are equal. Those matrices are given by the products of all weight matrices in the neural network $(\Wxf,1)$ for $\Psi(\Wxf,1)$, and the neural network $(\texttt{W}_{x'}^f,1)$ for $\Psi(\texttt{W}_{x'}^f,1)$. The final product of those two matrices can only be preserved by adding factors to the rows and columns of the internal matrices that get cancelled with the successive matrix multiplications. In other words, the final product of those two matrices is invariant only under a change of basis of the quiver representations (cf. Chapter 1 of \citep{Barot15}). This implies that the quiver representations $\Wxf$ and $\texttt{W}_{x'}^f$ are isomorphic via $\Gt$.
 
%\begin{Remark} \label{rmk:dataarchitecture}

\subsection{Consequences}
Interpreting data as quiver representations {\color{black}has} several consequences.

\paragraph{Consequence 1}
    The combinatorial architecture of $(W,f)$ and of $(\Wxf,1)$ are equal, and the weight architecture of $(\Wxf,1)$ is determined by both the weight and activation architectures of the neural network $(W,f)$ when its fed the input vector $x$. {\color{black} This means that even though the network function is non-linear because of the activation functions, all computations of the forward pass of a network on a given input vector can be arranged into a linear object (the quiver representation $\Wxf$), while preserving the output of the network, by Theorem~\ref{thm:data}. 
    
    {\color{black}Even more, feature maps and outputs of hidden neurons can be recovered completely from the quiver representations $\Wxf$, which implies that the notion \citep{Bengio13,Hinton07} of \textit{representation created by a neural network} in deep learning is a mathematical consequence of understanding data as quiver representations. }
    
    It is well known that feature maps can be visualized into images showing the input signal characteristics and thus providing intuitions on the behavior of the network and its impact on an image~\citep{Bengio13,Hinton07,Yosinski15,Feghahati19}. This notion is implied by our findings as our thin quiver representations of data $\Wxf$ include both the network structure and the feature maps induced by the data, expressed by the formula
     \[
        f_v\left( \act(\Wxf,1)_v(1^d) \right) = \act(W,f)_v(x),
     \]
     see Eq.~(\ref{Wxf}) in page \pageref{Wxf} and the proof of Theorem~\ref{thm:data}.  
    
    {\color{black}Practically speaking, it is useless to compute the quiver representation $\Wxf$ only to recover the outputs of hidden neurons, that are even more efficiently computed directly from the forward pass of data. Nevertheless, the way in which the outputs of hidden neurons are obtained from the quiver representations $\Wxf$ is by forgetting algebraic structure, more specifically forgetting pieces of the quiver, which is formalized by the notion of \textit{forgetful functors} in representation theory. All this implies that the notion of representation in deep learning is obtained from the quiver representations $\Wxf$ by loosing information of the computations of the neural network.}

 As such, using a thin quiver representation opens the door to a formal (and less intuitive) way to understand the interaction between data and the structure of a network, that takes into account all the combinatorics of the network and not only the activation outputs of the neurons, as it is currently understood}. %This adds up to the knowledge on how a neural network creates hierarchical representations from data \citep{Bengio13,Hinton07}, that takes into account the combinatorial and algebraic properties of neural networks.}
   
\paragraph{Consequence 2}
% questions for future research
% possible outcomes
% future outcomes
% induced lines of inquiry

    {\color{black}
    \begin{Corollary} \label{cor:wxfisom}
        Let $(x,t)$ and $(x',t')$ be data samples for $(W,f)$. If the quiver representations $\Wxf$ and $\texttt{W}_{x'}^f$ are isomorphic via $\Gt$ then $\Psi(W,f)(x) = \Psi(W,f)(x')$.
    \end{Corollary}
    {\bf Proof}.
    The neural networks $(\Wxf,1)$ and $(\texttt{W}_{x'}^f,1)$ are isomorphic if and only if the quiver representations $\Wxf$ and $\texttt{W}_{x'}^f$ are isomorphic via $\Gt$. By the last Theorem and the fact that isomorphic neural networks have the same network function (Theorem \ref{thm:netfunc}) we get that
    \[
        \Psi(W,f)(x) = \Psi(\Wxf,1)(1^d) = \Psi(\texttt{W}_{x'}^f,1)(1^d) = \Psi(W,f)(x').
    \]\hfill\BlackBox
    
    By this Corollary and the invariance of the network function under isomorphisms of the group $\Gt$ (Theorem~\ref{thm:netfunc})}, we obtain that the neural network is representing the data and the output on $(W,f)$ as the isomorphism classes $\left[ \Wxf \right]:= \{ \tau \cdot \Wxf : \tau \in \Gt \}$ of the thin quiver representations $\Wxf$ under the action of the change of basis group $\Gt$ of neural networks. {\color{black}This motivates the construction of a space whose points are isomorphism classes of quiver representations, which is exactly the construction of ``\textit{moduli space}'' presented in the next section.}  
    
\subsection{Induced inquiry for future research}

{\color{black}The language of quiver representations applied to neural networks brings new perspectives on their behavior and thus is likely to open doors for future works.  Here is one inquiry for the future.}

%    \paragraph{Inquiry 1}
    If a data sample $x$ is represented by a thin quiver representation $\Wxf$, one can generate an infinite amount of new data representations $\texttt{W}_{x'}^f$ via  $\Gt$ which all have the same network output, {\color{black} by applying an isomorphism given by $\tau \in \Gt$ using Eq.~(\ref{eq:group_action}), and then constructing an input $x'$ from it that produces such isomorphic quiver representation}. Doing so could have important implications in the field of adversarial attacks and network fooling~\citep{Akhtar18} where one could generate fake data at will which, when fed to a network, all have exactly the same output than the original data $x$. {\color{black} This will require the construction of a map from quiver representations to the input space, which could be done by using tools from algebraic geometry to find sections of the map $x \mapsto \left[ \Wxf \right]$, for which the construction of the moduli space in the next section is necessary, but not sufficient. This leads us to propose the following question for future research:
    \[
        \text{``Can the data quiver representations } \Wxf \text{ be translated back to input data?.''}
    \]
    
    Following the same logic, one could use this for data augmentation.  Starting from an annotated dataset $D=\{(x_1,t_1), (x_2,t_2), ..., (x_N,t_N)\}$, one could represent each data $x_i$ by a thin quiver representation : $\texttt{W}_{x_i}^f$, apply an arbitrary number of isomorphisms to it: $\{\tau^1\texttt{W}_{x_i}^f, \tau^2\texttt{W}_{x_i}^f,...,\tau^M\texttt{W}_{x_i}^f\}$ and then convert these representations back to the input data space.
    }

%\end{Remark}
%\begin{Remark} \label{rmk:featurespace}

%    We can see that the definition of feature space~\citep{Yosinski15,Feghahati19} for neural networks in deep learning is compatible with our findings.For a given input vector $x$, the isomorphism classes $\left[ \Wxf \right]$ encode all the feature space coordinates at the same time, as it is shown in the proof of Theorem \ref{thm:data}. Namely that $f_v\left( \act(\Wxf,1)_v(1^d) \right) = \act(W,f)_v(x)$ for every $v \in \V$ hidden or output vertex. When the activation architecture of $(W,f)$ is ReLU, either $\act(W,f)_v(x)=0$ or $\act(W,f)_v(x)= \act(\Wxf,1)_v(1^d)$ for every $v \in \V$ hidden or output vertex, this can also be seen in Appendix~\ref{appendix2}.

    \section{The Moduli Space of a Neural Network} \label{sec:moduli}

In this section, we propose a modified version of the manifold hypothesis~\citet[section~5.11.3]{Goodfellow-et-al-2016}. The original manifold hypothesis claims that the data lies in a small dimensional manifold inside the input space. {\color{black}We will provide an explicit map from the input space to the moduli space of a neural network with which the data manifold can be translated to the moduli space. This will allow the use of mathematical theory for quiver moduli spaces \citep{Das19,Reineke08,Reineke20} to manifold learning, representation learning and the dynamics of neural network learning \citep{Bengio13,Hinton07}.}

%Although this hypothesis seems intuitive and the empirical findings support its formulation, the existence and the structure of such manifold is mathematically vague.  Our manifold hypothesis will put more geometric structure on the manifold formed by algebraic objects, and we will give an explicit map whose image generates a manifold containing the data quiver representations of a data set.

{\color{black}
\begin{Remark}
    Throughout this section, we assume that all the weights of a neural network and of the induced data representations $\Wxf$ are non-zero. This can be assumed since the set where some of the weights are zero is of measure zero, and even in the case where it is exactly zero we can add a small number to it to make it non-zero and at the same time imperceptible to the computations of any computer, for example, infinitesimally smaller than the machine epsilon.
\end{Remark}
}

In order to formalize our manifold hypothesis, we will attach an explicit geometrical object to every neural network $(W,f)$ over a network quiver $Q$, that will contain the isomorphism classes of the data quiver representations $\left[ \Wxf \right]$ induced by any kind of data set $D$. This geometrical object that we denote $_d\mathcal{M}_k(\Q)$ is called the \textbf{moduli space}.  The moduli space only depends on the combinatorial architecture of the neural network, while the activation and weight architectures of the neural network determine how the isomorphism classes of the data quiver representations $\left[\Wxf \right]$ are distributed inside the moduli space. 

%We are interested in the following finite set
%\[
%    \mathcal{N} \Big( (W,f), D \Big) := \left\{ \left[ \Wxf \right] : (x,y) \in D \right\},
%\]
%where $\left[ \Wxf \right] = \{ \tau \cdot \Wxf : \tau \in \Gt \}$, and $\Gt$ is the change of basis group of neural networks.  In the following Theorem we show that the set of isomorphism classes of data quiver representations is invariant under isomorphisms of neural networks.
%\begin{Theorem} \label{thm:isominvsvarphi}
%    Let $(W,f)$ and $(V,g)$ be neural networks over $Q$ and let $\tau:(W,f) \to (V,g)$ be an isomorphism of neural networks. Then
%    \[
%        \mathcal{N} \Big( (W,f), D \Big) = \mathcal{N} \Big( (V,g), D \Big),
%    \]
%    for all dataset $D$ for $Q$.
%\end{Theorem}
% {\bf Proof}.
%   We will show that for any sample $(x,t) \in D$ we have that $\left[ \Wxf \right] = \left[ \Vxg \right]$, and therefore both sets $\mathcal{N} \Big( (W,f), D \Big)$ and $\mathcal{N} \Big( (V,g), D \Big)$ are equal. Since $\tau:(W,f) \to (V,g)$ is an isomorphism of neural networks, by Theorem~(\ref{thm:netfunc}) we have that
%    \[
%        \Psi(\Wxf,1)(1^d) = \Psi(W,f)(x) = \Psi(V,g)(x) = \Psi(\Vxg,1)(1^d),
%    \]
%    thus $\Wxf$ and $\Vxg$ are isomorphic via $\Gt$, as in the proof of Corollary \ref{cor:wxfisom}. Therefore $\left[\Wxf\right]=\left[\Vxg \right]$, which finishes the proof.
% 
The mathematical objects required to formalize our manifold hypothesis are known as \textbf{framed quiver representations}.  We will follow~\citet{Reineke08} for the construction of framed quiver representations in our particular case of thin representations.  Recall that the \textbf{hidden quiver} $\Q = (\widetilde{\V}, \widetilde{\mathcal{E}}, \widetilde{s}, \widetilde{t})$ of a network quiver $Q$ is the sub-quiver of the delooped quiver $Q^\circ$ formed by the hidden vertices $\VV$ and the oriented edges $\widetilde{\mathcal{E}}$ between hidden vertices. Every thin representation of the delooped quiver $Q^\circ$ induces a thin representation of the hidden quiver $\Q$ by forgetting the oriented edges whose source is an input (or bias) vertex, or the target is an output vertex. 

{\color{black}
\begin{Definition}
    We call \textbf{input vertices} of $\Q$ the vertices of $\Q$ that are connected to the input vertices of $Q$, and we call \textbf{output vertices} of $\Q$ the vertices that are connected to the output vertices of $Q$. 
\end{Definition}

Observe that the input vertices of the hidden quiver $\Q$ may not all of them be source vertices, so in the neural network we allow oriented edges from the input layer to deeper layers in the network. Dually, the output vertices of the hidden quiver $\Q$ may not all of them be sink vertices, so in the neural network we allow oriented edges from any layer to the output layer.}  %Let $d_1$ be the number of input vertices of $\Q$ and let $d_L$ be the number of output vertices of $\Q$. That is, there are $d_1$ neurons in the first hidden layer, and $d_L$ neurons in the last hidden layer.

\begin{Remark}
    For the sake of simplicity, we will assume that there are no bias vertices in the quiver $Q$. If there are bias vertices in $Q$, we can consider them as part of the input layer in such a way that every input vector $x \in \Cmplx^d$ needs to be extended to a vector $x' \in \Cmplx^{d+b}$ with its last $b$ coordinates all equal to 1, where $b$ is the number of bias vertices.  All the quiver representation theoretic arguments made in this section are therefore valid also for neural networks with bias vertices under these considerations. This also has to do with the fact that the group of change of basis of neural networks $\Gt$ has no factor corresponding to bias vertices, as the hidden quiver is obtained by removing all source vertices, not only input vertices.
\end{Remark}

Let $\widetilde{W}$ be a thin representation of $\Q$. We fix once and for all a family of vector spaces $\{ V_v \}_{v \in \VV}$ indexed by the vertices of $\Q$, given by $V_v=\Cmplx^k$ when $v$ is an output vertex of $\Q$ and $V_v=0$ for any other $v\in \VV$.
{\color{black}
\begin{Definition}~\citep{Reineke08}
    A choice of a thin representation $\widetilde{W}$ of the hidden quiver and a map $h_v:\widetilde{W}_v \to V_v$ for each $v \in \VV$ determines a pair $(\widetilde{W},h)$, where $h=\{ h_v \}_{v \in \VV}$, that is known as a \textbf{framed} quiver representation of $\Q$ by the family of vector spaces $\{ V_v \}_{v \in \VV}$.
\end{Definition}
}
We can see that $h_v$ is equal to the zero map when $v$ is not an output vertex of $\Q$, and %{\color{black} each $h_v$  } $h=(h_v)_{v \in \VV}$ can be represented by a matrix from the output layer of $\widetilde{W}$ to $\Cmplx^k$, so that we can omit the vector spaces $V_v$ from the notation of $h$. 
$h_v : \Cmplx \to \Cmplx^k$ for every $v$ output vertex of $\Q$.% and if $h$ is non-zero then there is at least one $h_v$ that is non-zero for $v$ an output vertex of $\Q$.

Dually, we can fix a family of vector spaces $\{ U_v \}_{v \in \VV}$ indexed by $\VV$ and given by $U_v=\Cmplx^d$ when $v$ is an input vertex of $\Q$ and $U_v=0$ for any other $v \in \VV$. 
{\color{black}
\begin{Definition}~\citep{Reineke08}
    A choice of a thin representation $\widetilde{W}$ of the hidden quiver and a map $\ell_v:U_v \to \widetilde{W}_v$ for each $v\in \VV$ determines a pair $(\widetilde{W},\ell)$, where $\ell=\{ \ell_v \}_{v \in \VV}$, that is known as a \textbf{co-framed} quiver representation of $\Q$ by the family of vector spaces $\{ U_v \}_{v \in \VV}$.
\end{Definition}
}

We can see that $\ell_v$ is the zero map when $v$ is not an input vertex of $\Q$, and %therefore $\ell=(\ell_v)_{v \in \VV}$ can be represented by a matrix from $\Cmplx^d$ to the input layer of $\widetilde{W}$, so that we can omit the vector spaces $U_v$ from the notation of $\ell$. Observe that 
$\ell_v: \Cmplx^d \to \Cmplx$ for every $v$ an input vertex of $\Q$. % and if $\ell$ is non-zero then there is at least one $\ell_v$ that is non-zero for $v \in \VV$ an input vertex of $\Q$. If we were to consider bias vertices in $Q$, the weights corresponding to the bias vertices will form part of $\ell$, and $\ell$ will no longer be represented by a matrix, but of course the quiver representation arguments will still hold.

%We will consider both $h$ and $\ell$ as matrices with coordinates $h_{i,j}$ and $\ell_{i,j}$ respectively.
%\begin{Remark}
%    We will make both the framing and the coframing at the same time to thin representations of $\Q$. This is compatible with the results of \citep{Reineke08}.
%\end{Remark}
\begin{Definition}
    A \textbf{double-framed} thin quiver representation is a triple $(\ell,\widetilde{W},h)$ where $\widetilde{W}$ is a thin quiver representation of {\color{black}the hidden quiver, $(\widetilde{W},h)$ is a framed representation of $\Q$ and $(\widetilde{W},\ell)$ is a co-framed representation of $\Q$.} %$\ell: \Cmplx^d \to \Cmplx^{d_1}$ and $h:\Cmplx^{d_L} \to \Cmplx^k$ are non-zero linear maps (see Fig.~\ref{fig:Double_framed}). We let $_d\mathcal{R}_k(\Q)$ be the set of double-framed thin quiver representations of $\widetilde{Q}$.
\end{Definition}

\begin{Remark}
    In representation theory, one does either a framing or a co-framing, and chooses a stability condition for each one. In our case, we will do both at the same time, and use the definition of stability given by~\citep{Reineke08} for framed representations, together with its dual notion of stability for co-framed representations.
\end{Remark}

\begin{figure} 
\centering
%\begin{subfigure}{.5\textwidth}
%  \centering
  \includegraphics[width=.5\linewidth]{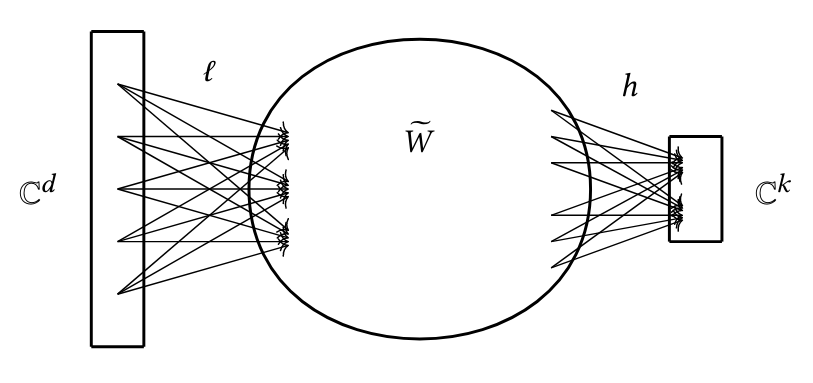}
\caption{An illustration of a double-framed thin quiver representation $(\ell,\widetilde{W},h)$. The boxes define the vector spaces of the framing and co-framing, given by $\Cmplx^d$ and $\Cmplx^k$, respectively.% Each input vertex of $\Q$ is connected to $\Cmplx^d$ and each output vertex of $\Q$ is connected to $\Cmplx^k$.
} \label{fig:Double_framed}
\end{figure}

%\ctikzfig{doubleframed}
{\color{black}
\begin{Definition}
    The group of \textbf{change of basis of double-framed thin quiver representations} is the same group $\Gt$ of change of basis of neural networks.
\end{Definition}
}

The action of $\Gt$ on double-framed quiver representations for $\tau \in \Gt$ is given by
\[
    \tau \cdot (\ell,\widetilde{W},h) = ( \tau  \cdot \ell,\tau \cdot \widetilde{W}, \tau \cdot h),
\]
where {\color{black} each component of $\tau \cdot h$ is given by $(\tau \cdot h)_v := (h_v^1/ \tau_v,...,h_v^k/ \tau_v)$, if we express $h_v=(h_v^1,...,h_v^k)$, and each component of $\tau \cdot \ell$ is given by $(\tau \cdot \ell)_v := (\ell_v^1 \tau_v,...,\ell_v^k \tau_v)$, if we express $\ell_v=(\ell_v^1,...,\ell_v^k)$.} %the matrices $\tau \cdot h$ and $\tau \cdot \ell$ are given by $(\tau \cdot h)_{i,j} = h_{i,j} / \tau_i$ and $(\tau \cdot \ell)_{i,j} = \ell_{i,j} \tau_j$. 
Every double-framed thin quiver representation of $\Q$ isomorphic to $(\ell,\widetilde{W},h)$ is of the form $\tau \cdot (\ell,\widetilde{W},h)$ for some $\tau \in \Gt$.
In the following theorem, we show that instead of studying the isomorphism classes $\left[ \Wxf \right]$ of the thin quiver representations of the delooped quiver $Q^\circ$ induced by the data, we can study the isomorphism classes of double-framed thin quiver representations of the hidden quiver.
\begin{Theorem} \label{thm:bijectivecorrespondence}
    There exists a bijective correspondence between the set of isomorphism classes $[W]$ via $\Gt$ of thin representations over the delooped quiver $Q^\circ$ and the set of isomorphism classes $[(\ell, \widetilde{W}, h)]$ of double-framed thin quiver representations of $\Q$.
\end{Theorem}
 {\bf Proof}.
    The correspondence between isomorphism classes is due to the equality of the group of change of basis for neural networks and double-framed thin quiver representations, since the isomorphism classes are given by the action of the same group.
    Given a thin representation $W$ of the delooped quiver, it induces a thin representation $\widetilde{W}$ of the hidden quiver $\Q$ by forgetting the input and output layers of $Q$. Moreover, if we consider the input vertices of $Q$ as the coordinates of $\Cmplx^d$ and the output vertices of $Q$ as the coordinates of $\Cmplx^k$, then the weights starting on input vertices of $Q$ define the map $\ell$ while the weights ending on output vertices of $Q$ define the map $h$. This can be seen in Fig.~\ref{fig:Double_framed}. Given a double-framed thin quiver representation $(\ell,\widetilde{W},h)$, the entries $\ell$  (resp. $h$) are the weights of a thin representation $W$ starting (resp. ending) on input (resp. output) vertices, while $\widetilde{W}$ defines the hidden weights of $W$.\hfill\BlackBox

From now on we will identify a double-framed thin quiver representation $(\ell,\widetilde{W},h)$ with the thin representation $W$ of the delooped quiver $Q^\circ$ defined by $(\ell,\widetilde{W},h)$ as in the proof of the last theorem. We will also identify the isomorphism classes
\[
    \left[W\right] = \left[(\ell,\widetilde{W},h)\right],
\]
where the symbol on the left means the isomorphism class of the thin representation $W$ under the action of $\Gt$, and the one on the right is the isomorphism class of the double-framed thin quiver representation $(\ell,\widetilde{W},h)$.

{\color{black}
One would like to study the space of \textit{all} isomorphism classes of double-framed thin representations of the delooped quiver. However, it is well-known that this space does not has a good topology \citep{Nakajima94}. Therefore, one considers the space of isomorphism classes of \textit{stable} double-framed thin quiver representations instead of all quiver representations, that can be shown to have a much richer topological and geometrical structure. In order to be stable, a representation has to satisfy a stability condition that is given in terms of its \textit{sub-representations}. We will prove that the data representations $\Wxf$ are stable in this sense, and to do so we will now introduce the necessary definitions.
}

%Before using~\citet{Nakajima94}'s theorem on the existence of the moduli space, and~\citet{Reineke08}'s calculation of the dimension of the moduli space, we will need to consider sub-representations of a quiver representation.

\begin{Definition} \citep[page~14]{Schiffler14} \label{def:sub_rep}
    Let $W$ be a thin representation of the delooped quiver $Q^\circ$ of a network quiver $Q$. A \textbf{sub-representation} of $W$ is a representation $U$ of $Q^\circ$ such that there is a morphism of representations $\tau:U \to W$ where each map $\tau_v$ is an injective map.
    %for each $v\in \V$ we have that $dim(U_v) \leq dim(W_v)$ and there exists a non-zero morphism of representations $\tau:U \to W$. %If the morphism $\tau$ is non-zero and $U_v\not=0$ for at least one $v\in \V$, we say that the sub-representation $U$ is a \textbf{proper sub-representation} of $W$.
\end{Definition}
\begin{figure} 
\centering
%\begin{subfigure}{.5\textwidth}
%  \centering
(a)  \includegraphics[width=.25\linewidth]{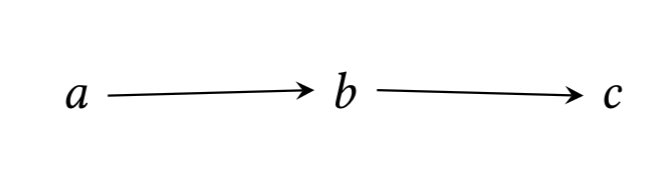}
(b)  \includegraphics[width=.3\linewidth]{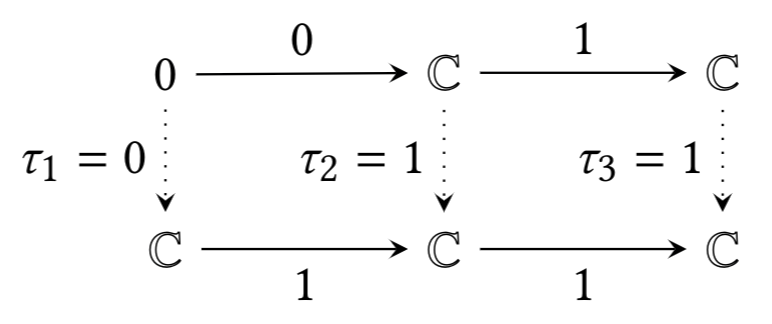}
(c)  \includegraphics[width=.3\linewidth]{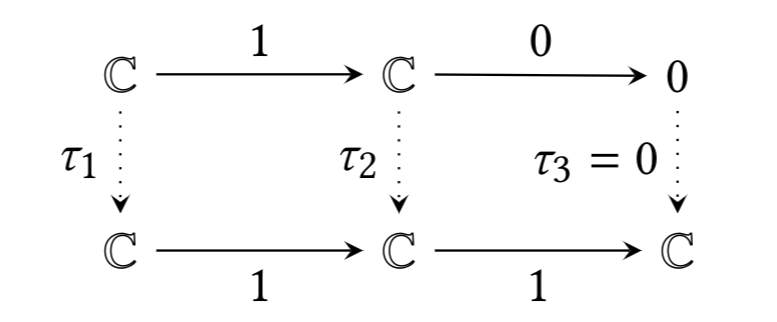}
\caption{(a) A quiver $Q$.  (b) A diagram showing a quiver representation of $Q$ at the bottom and a sub-representation at the top together with the morphism of representations $\tau$ from Definition~\ref{def:sub_rep} given by dotted oriented edges.  (c) A diagram showing a quiver representation of $Q$ at the bottom and another one at the top. The representation at the top is not a sub-representation of the bottom representation because there is no possible choice of morphism of representations $\tau$ that is injective in every vertex.} \label{fig:sub_representation}
\end{figure}

%For example, the following diagram shows a proper sub-representation (top) of a representation (bottom):
%\[
%	\begin{tikzpicture}
%         	\matrix (m) [matrix of math nodes,row sep=2em,column sep=2em]
%          	{
%           	  	 a & & b & & c \\  
%           	};
%          	\path[-stealth]
%      	 (m-1-1) edge node [above] {} (m-1-3)
%      	 (m-1-3) edge node [above] {} (m-1-5);
%	\end{tikzpicture}
%\]
%Note that in the following diagram, there is no possible choice of non-zero $\tau$ making it commutative, so the top row does not defines a sub-representation:
%\[
%	\begin{tikzpicture}
%         	\matrix (m) [matrix of math nodes,row sep=2em,column sep=2em]
%          	{
%           	  	 \Cmplx & & \Cmplx & & 0 \\ 
%           	  	 \Cmplx & & \Cmplx & & \Cmplx \\ 
%           	};
%          	\path[-stealth]
%      	 (m-1-1) edge node [above] {1} (m-1-3)
%      	 (m-1-3) edge node [above] {0} (m-1-5)
%      	 (m-2-1) edge node [below] {1} (m-2-3)
%      	 (m-2-3) edge node [below] {1} (m-2-5)
      	 
%      	 (m-1-1) edge[dotted] node [left] {$\tau_1$} (m-2-1)
%      	 (m-1-3) edge[dotted] node [left] {$\tau_2$} (m-2-3)
%      	 (m-1-5) edge[dotted] node [left] {$\tau_3=0$} (m-2-5);
%	\end{tikzpicture}
%\]
{\color{black}
\begin{Definition}
    The \textbf{zero representation} of $Q$ is the representation denoted $0$ where every vector space assigned to every vertex is the zero vector space, and therefore every linear map in it is also zero.
\end{Definition}
}

Note that if $U$ is a quiver representation, then the zero representation $0$ is a sub-representation of $U$ since $\tau_v=0$ is an injective map {\color{black}in this case}.

%For instance, a representation $U$ such that $U_v=0$ on one single output vertex $v$ cannot be a sub-representation of a thin representation unless $U_v=0$ for all $v$.  This fact will be used in the proof of the following theorem, together with its dual notion.  Namely, the only way a representation $U$ can be a sub-representation of a thin representation with $U_v= \Cmplx$ for at least one input vertex $v$ is when $U_v=\Cmplx$ for all vertices $v$. For example, in Fig.~\ref{fig:sub_representation}(c) we can see that the top representation is not a sub-representation of the bottom thin representation. %But if the vector space in the sink of the quiver was $\Cmplx$, then it will become a sub-representation that is actually isomorphic to the bottom thin representation.
%\begin{Definition} \citep{Schiffler14}

%The existence of the moduli space depends on a chosen notion of \textit{stability} for double-framed quiver representations~\citep{Reineke08,Nakajima94}.  We will introduce this notion and then prove that every double-framed thin quiver representation induced by a neural network and a data sample is stable. 

We can see from Fig.~\ref{fig:sub_representation} that the combinatorics of the quiver are related to the existence of sub-representations. {\color{black}Therefore, we explain now how to use the combinatorics of the quiver to prove stability of our data-representations $\Wxf$.}

Given a double-framed thin quiver representation $(\ell,\widetilde{W},h)$, the image of the map $\ell$ lies inside the representation $\widetilde{W}$. The map $\ell$ is given by a family of maps indexed by the vertices of the hidden quiver $\Q$, namely $\ell = \{ \ell_v : \Cmplx^{n_v} \to \widetilde{W}_v \ | \ v \in \VV \}$. Recall that $n_v=0$ if $v$ is not an input vertex of the hidden quiver $\Q$, and $n_v=d$ when $v$ is an input vertex of $\Q$.  The image of $\ell$ is by definition a family of vector spaces indexed by the hidden quiver $\Q$, given by
\[
    Im(\ell) = \big( Im(\ell)_v \big)_{v \in \VV} \text{  where  } Im(\ell)_v \subset \widetilde{W}_v.
\]
By definition $Im(\ell)_v = \{ z \in \widetilde{W}_v = \Cmplx \ | \ \ell_v(w)=z \text{ for some } w\in \Cmplx^{n_v} \}$. {\color{black}Recall that we will interpret the data quiver representations $\Wxf$ as double-framed representations, and that $\Wxf$ respects the output of the network $(W,f)$ when it is fed the input vector $x$. According to Eq.~(\ref{Wxf}), the weights in the input layer of $\Wxf$ are given in terms of the weights in the input layer of the network $W$ and the input vector $x$. Therefore, only on a set of measure zero we have that some of the weights in the input layer of $\Wxf$ are zero, so we can assume, without loss of generality, that the weights on the input layer of $\Wxf$ are all non-zero.}

%As a consequence, we can assume there is at least one map $\ell_v$ that is non-zero in the output layer of $\Wxf$, otherwise the output of the network will be a vector of only zeros, which in reality is an event of probability zero.
% and therefore $Im(\ell)_v$ is non-zero for at least one input vertex $v$ of $\Q$. % zero whenever $v$ is not an input vertex of $\Q$. %And if $\ell_v$ is not the zero map, then $Im(\ell)_v = \Cmplx$.

Dually, the kernel of the map $h$ lies inside the representation $\widetilde{W}$. The map $h$ is given by a family of maps indexed by the vertices of the hidden quiver $\Q$, namely $h = \{ h_v :\widetilde{W}_v \to  \Cmplx^{m_v} \ | \ v \in \VV \}$. Recall that $m_v=0$ if $v$ is not an output vertex of the hidden quiver $\Q$, and $m_v=k$ when $v$ is an output vertex of $\Q$.  Therefore, the kernel of $h$ is by definition a family of vector spaces indexed by the hidden quiver $\Q$. That is,
\[
    ker(h) = \big( ker(h)_v \big)_{v \in \VV} \text{  where  } ker(h)_v \subset \widetilde{W}_v.
\]
By definition $ker(h)_v = \{ z \in \widetilde{W}_v \ | \ h_v(z)=0 \}$. {\color{black}The set where all of $h_v$ are equal to zero is of measure zero, and even in the case where it is exactly zero we can add a very small number to every coordinate of $h_v$ to make it non-zero and that the output of the network doesn't significantly changes. So we can assume, without loss of generality, that all the maps $h_v$ are non-zero for every output vertex $v$ of $\Q$.}

 % and therefore $ker(h)_v$ is zero for at leas one output vertex $v$ of $\Q$.
%whenever $v$ is not an output vertex of $\Q$. And if $h_v$ is not the zero map, then $ker(h)_v = 0$ the zero vector space.
%\end{Definition}
\begin{Definition} \label{def:stable}
    A double-framed thin quiver representation $(\ell,\widetilde{W},h)$ is \textbf{stable} if the following two conditions are satisfied:
    \begin{itemize}
        \item[1.] The only sub-representation $U$ of $\widetilde{W}$ which is contained in $ker(h)$ is the zero sub-representation, and
        \item[2.] The only sub-representation $U$ of $\widetilde{W}$ that contains $Im(\ell)$ is $\widetilde{W}$.
    \end{itemize}
\end{Definition}
\begin{Theorem}
    Let $(W,f)$ be a neural network and let $(x,t)$ be a data sample for $(W,f)$. Then the double-framed thin quiver representation $\Wxf$ is stable.
\end{Theorem}
 {\bf Proof}.
    We express $\Wxf=(\ell,\widetilde{W},h)$ as in Theorem \ref{thm:bijectivecorrespondence}. {\color{black}As explained before Definition~\ref{def:stable}, we can assume, without loss of generality, that for every input vertex $v$ of $\Q$ the map $\ell_v$ is non-zero, and that for every output vertex $v$ of $\Q$ the map $h_v$ is non-zero.
    
    %there is at least one map $h_v$ that is non-zero, and that there is at least one map $\ell_v$ that is non-zero, {\color{black}as explained before }
    
    {\color{black}We have that $h_v:\Cmplx \to \Cmplx^k$ is a linear map, so its kernel is either $0$ or $\Cmplx$. But $Ker(h_v)=\Cmplx$ if and only if $h_v=0$, and since $h_v\not=0$ we get that $Ker(h_v)=0$}
    %The kernel $ker(h)$ is zero in at least one output vertex of $\Q$ 
    and, as in Fig.~\ref{fig:sub_representation}, after the combinatorics of quiver representations, there is no sub-representation of $\widetilde{W}$ with all its factors corresponding to output vertices of $\Q$, other than the zero representation. Since the combinatorics of network quivers forces a sub-representation contained in $ker(h)$ to be the zero sub-representation, we obtain the first condition for stability of double-framed thin quiver representations.
    
    Dually, we have that $\ell_v:\Cmplx^d \to \Cmplx$ is a linear map, so its image is either $0$ or $\Cmplx$. But $Im(\ell_v)=0$ if and only if $\ell_v=0$, and since $\ell_v\not=0$ we get that $Im(\ell_v)=\Cmplx$ and,}
    %the image $Im(\ell)_v$ is equal to $\widetilde{W}_v = \Cmplx$ for at least one input vertex $v \in \VV$.  
    as in Fig.~\ref{fig:sub_representation}, there is no sub-representation of $\widetilde{W}$ that contains $Im(\ell)$ other than $\widetilde{W}$. Therefore the only sub-representation of $\widetilde{W}$ that contains $Im(\ell)$ is $\widetilde{W}$. 
    
    Thus, $\Wxf=(\ell,\widetilde{W},h)$ is a stable double-framed thin quiver representation of the hidden quiver $\Q$.\hfill\BlackBox
 
{\color{black}
Denote by $_d\mathcal{R}_k(\Q)$ the space of all double-framed thin quiver representations. 
\begin{Definition}
    The \textbf{moduli space} of stable double-framed thin quiver representations of $\Q$ is by definition
\[
    _d\mathcal{M}_k(\Q) := \big\{ [V] : V \in  \ _d\mathcal{R}_k(\Q) \text{ is stable} \big\}.
\]
\end{Definition}
Note that the moduli space depends on the hidden quiver $\Q$ and the chosen vector spaces from which one double-frames the thin representations.

Given a neural network $(W,f)$ and an input vector $x\in \Cmplx^d$, we can define a map
\[
    \begin{array}{cccc}
        \varphi(W,f) : & \Cmplx^d & \to & \ _d\mathcal{R}_k(\Q)  \\
         & x & \mapsto & \Wxf.
    \end{array}
\]
By the last theorem, in the case where all the weights of $\Wxf$ are non-zero, this map takes values in the moduli space which parametrizes isomorphism classes of stable double-framed thin quiver representations
\[
    \begin{array}{cccc}
        \varphi(W,f) : & \Cmplx^d & \to & \ _d\mathcal{M}_k(\Q).  \\
    \end{array}
\]

\begin{Remark}
    For ReLU activations one can produce representations with some weights $\left( \Wxf \right)_\epsilon =0$. But note that these representations $\Wxf$ can be arbitrarily approximated by representations with non-zero weights. Nevertheless, the map $\varphi(W,f)$ with values in $_d\mathcal{R}_k(\Q)$ still decomposes the network function as in Consequence 1 below.
\end{Remark}
%\begin{Remark}
%    Usually one denotes by $_d\mathcal{R}_k^s(\Q)$ the set of stable double-framed thin quiver representations of $\Q$. Since $_d\mathcal{R}_k(\Q) = _d\mathcal{R}_k^s(\Q)$ we will not be using this notation. Recall also that our notion of stability is not the same as the one considered in \citep{Chindris20}; and that we use quiver representations to represent the data in terms of the network, and not the data by itself.
%\end{Remark}

}

%Therefore, we have proved that given a neural network $(W,f)$ and a data set $D$, the isomorphism classes of the data quiver representations $\left[ \Wxf \right]$ lie inside the moduli space. This fact doesn't depend on the weights of the network, the activations, the chosen architecture, the training nor the chosen data set. It only depends on the quiver $Q$, that is, the combinatorial architecture of the network.

The following result is a particular case of~\citet{Nakajima94}'s theorem, generalized for double-framings and restricted to thin representations, combined with~\citet{Reineke08}'s calculation of framed quiver moduli space dimension adjusted for double-framings (see Appendix~\ref{appendix3} for details about the computation of this dimension).
\begin{Theorem} \label{thm:nakajima}
    Let $Q$ be a network quiver. There exists a {\color{black}geometric quotient} $_d\mathcal{M}_k(\Q)$ of by the action of the group $\Gt$, called the \textbf{moduli space} of stable double-framed thin quiver representations of $\Q$. Moreover, $_d\mathcal{M}_k(\Q)$ is non-empty and {\color{black}its complex dimension is}
    \[
        dim_\Cmplx \left( _d\mathcal{M}_k(\Q) \right) = \# \mathcal{E}^\circ - \# \VV.
    \]
    In short, the dimension of the moduli space of the hidden quiver $\Q$ equals the number of edges of $Q^\circ$ minus the number of hidden vertices.
\end{Theorem}
\begin{Remark}
    The mathematical existence of the moduli space~\citep{Reineke08,Nakajima94} depends on two things,
    \begin{itemize}
        \item {\color{black}the neural networks and the data may be build upon the real numbers, but we are considering them over the complex numbers}, and %the models are build upon the complex numbers $\Cmplx$, and
        \item the change of basis group of neural networks $\Gt$ is the change of basis group of thin quiver representations of $\Q$, which is a reductive group.
    \end{itemize}
    One may try to study instead the space whose points are isomorphism classes given by the action of the sub-group $H$ of the change of basis group $\Gt$, whose action preserves both the weight and the activation architectures. By doing so we obtain a group $H$ that is not reductive, which gets in the way of the construction, and therefore the existence, of the moduli space. This happens even in the case of ReLU activation.
\end{Remark}

{\color{black}Finally, let us underline that the map $\varphi(W,f)$ from the input space to the representation space (i) takes values in the moduli space when all weights of the representations $\Wxf$ are non-zero, and (ii) may or may not be 1-1. And even if $\varphi(W,f)$ is not 1-1, all the results in this work still hold. The most important implication of the existence of the map $\varphi(W,f)$ is our Consequence 1 below, which does not depend on $\varphi(W,f)$ being 1-1.}

\subsection{Consequences}
\label{sec:muduli_consequences}

%\begin{Remark}

%\end{Remark}
{\color{black}
The existence of the moduli space of a neural network has the following consequences.
\paragraph{Consequence 1}
The moduli space $_d\mathcal{M}_k(\Q)$ as a set is given by
\[
    _d\mathcal{M}_k(\Q) = \left\{ [V] : V \in \ _d\mathcal{R}_k(\Q) \text{ is stable} \right\}.
\]
That is, the points of the moduli space are the isomorphism classes of (stable) double-framed thin quiver representations of $\Q$ over the action of the change of basis group $\Gt$ of neural networks. Given any point in the moduli space $[V]$ we can define 
\[
\hat{\Psi}[V]:=\Psi(V,1)(1^d)
\]
since the network function is invariant under isomorphisms, which gives a map 
$$\hat{\Psi}:_d\mathcal{M}_k(\Q) \to \Cmplx^k.$$ 
Also, given a neural network $(W,f)$, we define a map $\varphi(W,f) : \Cmplx^d \to _d\mathcal{M}_k(\Q)$ by
\[
    \varphi(W,f)(x) := \left[ \Wxf \right] \in \ \ _d\mathcal{M}_k(\Q).
\]
\begin{Corollary} \label{cor:netfunction}
    The network function of any neural network $(W,f)$ is decomposed as
    \[
    \Psi(W,f) = \hat{\Psi} \circ \varphi(W,f).
    \]
\end{Corollary}
{\bf Proof}. This is a consequence of Theorem~\ref{thm:data} since for any $x\in \Cmplx$ we have
\[
    \hat{\Psi} \circ \varphi(W,f) (x) = \hat{\Psi} \left[\Wxf\right] = \Psi(\Wxf,1)(1^d) =  \Psi(W,f)(x)._{\hfill\BlackBox}
\]
This implies that any decision of any neural network passes through the moduli space (and the representation space), and this fact is independent of the architecture, the activation function, the data and the task.
%Therefore, for every data sample $(x,t)$ the neural network $(W,f)$ induces a point $\left[\Wxf\right]$ in $_d\mathcal{M}_k(\Q)$. 
}
\paragraph{Consequence 2} \label{cons:7:4}
    Let $(W,f)$ be a neural network over $Q$ and let $(x,t)$ be a data sample. If $\left( \Wxf \right)_\epsilon=0$, then any other quiver representation $V$ of the delooped quiver $Q^\circ$ that is isomorphic to $\Wxf$ has $V_\epsilon=0$.
    Therefore, if in a dataset $\{ (x_i,t_i) \}_{i=1}^N$ {\color{black}the majority of} samples $(x,t)$ such that for a specific edge $\epsilon \in Q^\circ$ the corresponding weight on $\Wxf$ is zero, then the coordinates of $\left[\Wxf\right]$ inside the moduli space corresponding to $\epsilon$ are not used for computations. Therefore, a projection of those coordinates to zero corresponds to the notion of pruning of neural networks, that is forcing to zero the smaller weights on a network~\citep{Frankle18}.  From Eq.~(\ref{Wxf}) {\color{black}in page \pageref{Wxf}}, we can see that this interpretation of the data explains why naive pruning works. {\color{black}Namely, if one of the weights in the neural network $(W,f)$ is small, then so does the corresponding weight in $\Wxf$ for any input $x$. Since the coordinates of $\Wxf$ are given in function of the weights of $(W,f)$, by Eq. (5) in page 23 and the previous consequence, a small weight of $(W,f)$ sends inputs $x$ to representations $\Wxf$ with some coordinates equal to zero in the moduli space. If this happens for a big proportion of the samples in the dataset, then the network $(W,f)$ is not using all of the coordinates in the moduli space to represent its data in the form of the map $\varphi(W,f): \Cmplx^d \to _d\mathcal{M}_k(\Q)$. }
%\end{Remark}

{\color{black}
\paragraph{Consequence 3}

%It is usual to consider the network function of a neural network as the composition of the maps defined by its layers that produce the feature maps. Researchers have been putting attention to the patterns appearing in the higher layers of neural networks and how small changes in the previous layers can produce unexpected outputs, {\color{black} which is the case of adversarial examples~\citep{Akhtar18}}. Note that the quiver representation language allows to manage all .... without a mathematical footing that can manage all feature maps from early and deeper layers in the network at the same time, and this can be achieved by the representations $\Wxf$ and the moduli space in the following way.

Let $\mathcal{M}$ be the data manifold in the input space of a neural network $(W,f)$.  The map $\varphi(W,f)$ takes $\mathcal{M} \subset \Cmplx^d$ to $\varphi(W,f)\left( \mathcal{M} \right) \subset \ _d\mathcal{M}_k(\Q)$. The subset $\varphi(W,f)\left( \mathcal{M} \right)$ generates a sub-manifold of the moduli space (as it is well known in topology \citep{Munkres}) that parametrizes all possible outputs that the neural network $(W,f)$ can produce from inputs on the data manifold $\mathcal{M}$. This means that the geometry of the data manifold $\mathcal{M}$ has been translated into the moduli space $_d\mathcal{M}_k(\Q)$, and this implies that the mathematical knowledge \citep{Das19,Reineke08,Reineke20} that we have of the geometry of the moduli spaces $_d\mathcal{M}_k(\Q)$ can be used to understand the dynamics of neural network training, due to the universality of the description of neural networks we have provided.
}

\subsection{Induced inquiries for future research}

\paragraph{Inquiry 1}
Following Consequence 1, one would like to look for correlations between the dimension of the moduli space and properties of neural networks. The dimension of the moduli space is equal to the number of basis paths in ReLU networks found by~\citet{Zheng19}, where they empirically confirm that it is a good measure for generalization. This number was also obtained as the rank of a structure matrix for paths in a ReLU network~\citep{Meng18}, however, they put restrictions on the architecture of the network to compute it. {\color{black}As we noted before, the network function of any neural network passes through the moduli space, where the data quiver representations $\Wxf$ lie, so the dimension of the moduli space could be used to quantify the capacity of neural networks in general.}

\paragraph{Inquiry 2}
We can use the moduli space to formulate what training does to the data quiver representations.  Training a neural network through gradient descent generates an iterative sequence of neural networks $(W_1,f), (W_2,f), ... , (W_m,f)$ where $m$ is the total number of training iterations.  For each gradient descent iteration $i=1,...,m$ we have
\[
    Im \big( \varphi(W_i,f) \big) \ \ \subset \ \  _d\mathcal{M}_k(\Q).
\]
%Thus, the moduli space $_d\mathcal{M}_k(\Q)$ contains all possible isomorphism classes of double-framed thin representations given by any neural network with underlying quiver $Q$. So t
The moduli space is given only in terms of the combinatorial architecture of the neural network, while the weight and activation architectures determine how the points $\left[ \texttt{W}_{x_1}^f \right], ... , \left[ \texttt{W}_{x_n}^f \right]$ are distributed inside the moduli space $_d\mathcal{M}_k(\Q)$, {\color{black}because of Eq.~(\ref{Wxf})}. Since the training changes the weights and not (always) the network quiver (unless of course in neural architecture search), we obtain that each training step defines a different map $\varphi(W_i,f) : \Cmplx^d \to _d\mathcal{M}_k(\Q)${\color{black}. Therefore, the sub-manifold $Im \big( \varphi(W_i,f) \big)$ is changing its shape during training} inside the moduli space $_d\mathcal{M}_k(\Q)$.

A training of a neural network, which is a sequence of neural networks $(W_1,f),...,(W_m,f)$, can be thought as, first adjusting the manifold $Im \big( \varphi(W_1,f) \big)$ into $Im \big( \varphi(W_2,f) \big)$, then the manifold $Im \big( \varphi(W_2,f) \big)$ into $Im \big( \varphi(W_3,f) \big)$, and so on. This is a completely new way of representing the training of neural networks that works universally for any neural network, which leads to the question
{\color{black}
\begin{center}
    ``Can training dynamics be made more explicit in these moduli spaces in such a way that allows proving more precise convergence theorems than the currently known?''
\end{center}
}

\paragraph{Inquiry 3} \label{sec:7:cons}

A training of the form $(W_1,f),...,(W_m,f)$ only changes the weights of the neural network. As we can see, our data quiver representations depend on both the weights and the activations, and therefore a usual training does not exploits completely the fact that the data quiver representations are mapped via $\varphi$ to the moduli space. Thus, the idea of learning the activation functions, as it is done by~\citet{Goyal19}, will produce a training of the form $(W_1,f_1),...,(W_m,f_m)$, and this allows the maps $\varphi(W_i,f_i)$ to explore more freely the moduli space than the case where only the weights are learned. {\color{black}Our results imply that a training that changes (and not necessarily learns) the activation functions has the possibility of exploring more the moduli space due to the dependence of the map $\varphi(W,f)$ on the activation functions.  One would like to see if this can actually improve the training of neural networks, and these are exactly the results obtained by the experiments of \citet{Goyal19}. Therefore, the following question arises naturally.}
%Therefore, our results open up the door to a new way of training the weights and the activation functions of a neural network.
{\color{black}
\begin{center}
    ``Can neural network learning be improved by changing activation functions during training, for example with teleportation?''
\end{center}
}

\section{Conclusion and future works}

We presented the theoretical foundations for a different understanding of neural networks using their combinatorial and algebraic nature, while explaining current intuitions in deep learning by relying only on the mathematical consequences of the computations of the network during inference. We may summarize our work with the following five points,
\begin{itemize}
    \item[1.] We use quiver representation theory to represent neural networks and their data processing.
    \item[2.] This representation of neural networks scales to modern deep architectures like conv layers, pooling layers, residual layers, batch normalization {\color{black}and even randomly wired neural networks \cite{Xie19}}.
    \item[3.] Theorem~\ref{thm:netfunc} shows that neural networks are algebraic objects, in the sense that the maps preserving the algebraic structure also preserve the computations of the network. Even more, we show that positive scale invariance of ReLU networks is a particular case of this result.
    \item[4.] We represented data as thin quiver representations with identity activations in terms of the architecture of the network. We proved that this representation of data is algebraically consistent (invariant under isomorphisms) and carries the important notion of feature spaces of all layers at the same time.
    \item[5.] We introduced the moduli space of a neural network, and prove that it contains all possible (isomorphism classes of) thin quiver representations that result from the computations of the neural network on a forward pass.  This leads us to the mathematical formalization of a modified version of the manifold hypothesis in machine learning, given in terms of the architecture of the network.
    {\color{black}
    \item[6.] Our representation of neural networks and the data they process is the first to universally represent neural networks: it does not depends on the chosen architecture, activation functions, data, loss function, or even the task.
    }
\end{itemize}

To the knowledge of the authors, the insights, concepts and results in this work are the first of their kind.  In the future, we aim to translate more deep learning objects into the quiver representation language. For instance,

\begin{itemize}
    %\item Back propagation can be written in the quiver language. If $f'$ denotes the derivative of the activation functions of a neural network $(W,f)$, once the loss is computed, then $f_v\left(\act(\Wxf,1)_v(1^d)\right)$ together with $\act(W,f)_v(x)$, and of course $W$, are the factors used in back propagation to compute the gradient. Even more, the gradient gets encoded in a (double-framed thin) quiver representation that back propagates the error.
    \item Dropout~\citep{Srivastava14} is a restriction of the training to several network sub-quivers. This translates into adjustments of the configuration of the data inside the moduli space via sub-spaces given by sub-quivers.
    \item Generative adversarial networks~\citep{Goodfellow14} and actor-critics~\citep{Silver14} provide the stage for the interplay between two moduli spaces that get glued together to form a bigger one.
    %\item We believe that the latent space of autoencoders~\citep{Petscharnig17} is compatible with the moduli space. Transformations on the latent space correspond to transformations on a sub-space of the moduli space determined by the latent neurons.
    \item Recurrent neural networks~\citep{Hopfield88} become a stack of the same network quiver, and therefore the same moduli space gets glued with copies of itself multiple times.
    \item The knowledge stored in the moduli space in the form of the map $\varphi(W,f)$ provides a new concept to express and understand transfer learning~\citep{Baxter98}. Extending a trained network will globally change the moduli space, while fixing the map $\varphi(W,f)$ in the sub-space corresponding to the unchanged part of the network quiver.
\end{itemize}

On expanding the algebraic understanding of neural networks, we consider the following approaches for further research,
\begin{itemize}
    %\item Considering that the tensor product of thin representations is also a thin representation~\citep{Herschend08}, it becomes natural to study properties of (algebraic) tensor products of neural networks and of data representations.
    
    %\item Projective resolutions of thin representations can be computed and then we can use homological algebra~\citep{Assem06,Rotman08,Zimmermann14}.
    \item Study the possibility to transfer the gradient descent optimization to the moduli space with the goal of not only optimizing the network weights but also the activation functions.
    
    \item The combinatorics of network quivers seem key to the understanding of neural networks and their moduli spaces. So a further study of network quivers by themselves is required~\citep{Assem06,Barot15,Schiffler14}.
    
    \item Continuity and differentiability of the network function $\Psi(W,f)$ and the map $\varphi(W,f)$ will allow the use of more specific algebraic-geometric tools~\citep{Hartshorne77,Reineke08}. Even more, the moduli space is a toric variety and then we can use toric geometry~\citep{Cox11} to study the moduli space~\citep[see][]{Hille98,Domokos16}.
    
    \item Neural networks define finite-dimensional representations of wild hereditary finite-dimensional associative algebras, that can be studied with algebraic-combinatorial techniques~\citep{Assem06, delaPena94, Zimmermann14}.
\end{itemize}

Finally, this work provides a language in which to state a different kind of scientific hypotheses in deep learning, and we plan to use it as such. Many characterizations will arise from the interplay of algebraic methods and optimization.  Namely, when solving a task in practical deep learning one tries different hidden quivers and optimization hyperparameters.  Therefore, measuring changes in the hidden quiver will become important.

\acks{The first named author would like to thank Ibrahim Assem, Thomas Br{\"u}stle and Shiping Liu for the provided freedom of research. To Bernhard Keller and Markus Reineke for very useful mathematical exchange. We specially thank Thomas Br{\"u}stle and Markus Reineke for their help on computing the dimension of the moduli space provided in Appendix C. This paper was written while the first named author was a postdoctoral fellow at the Universit\'e de Sherbrooke. }


\newpage

\appendix

\section{Example of Theorem~\ref{thm:netfunc}} \label{appendix1}

Here we illustrate with an example the result of Theorem\ref{thm:netfunc}, i.e., that an isomorphism between two neural networks $(W,f)$ and $(V,g)$ preserves their network function $\Psi(W,f)(x) = \Psi(V,g)(x)$. Let's consider a ReLU multilayer perceptron $(W,f)$ with 2 hidden layers of 3 neurons each, 2 neurons on the input layer and 2 neurons on the output layer. That is,
\ctikzfig{mlp}
We denote by $W_1,W_2$ and $W_3$ the weight matrices of the network from left to right. Consider the explicit matrices
\[
    \begin{array}{ccc}
        W_1 = \left( \begin{array}{cc}
        0.2 & -0.4 \\
        -1.1 & 1.0 \\
        -0.1 & -0.2 \\
        \end{array} \right), & W_2 = \left( \begin{array}{ccc}
        -0.6 & -0.2 & -0.3 \\
        0.3 & 1.2 & -0.4 \\
        -0.1 & -1.0 & 0.2 \\
        \end{array} \right), & W_3 = \left( \begin{array}{ccc}
        0.5 & -0.7 & 0.3 \\
        -1.2 & 0.1 & -0.6 \\
        \end{array} \right). \\
    \end{array}
\]
Assume now that the input vector is the vector $x = \left( \begin{array}{c}
     -1.2  \\ 0.3 \\
\end{array} \right)$, then the output of the first layer is
\[
    ReLU(W_1 x) = ReLU \left( \begin{array}{c}
         0.2(-1.2) - 0.4(0.3)  \\
         -1.1(-1.2) + 1.0(0.3)  \\
         -0.1(-1.2) - 0.2(0.3)  \\
    \end{array} \right) = ReLU\left( \begin{array}{c}
         -0.36  \\ 1.62 \\ 0.06
    \end{array} \right) = \left( \begin{array}{c}
         0  \\ 1.62 \\ 0.06
    \end{array} \right)
\]
The output of the second layer is
\[
\begin{array}{lcl}
    ReLU \big( W_2(ReLU(W_1 x)) \big) & = & ReLU \left( \begin{array}{c}
         -0.2(1.62) - 0.3(0.06)  \\
         1.2(1.62) - 0.4(0.06)  \\
         -1.0(1.62) + 0.2(0.06)  \\
    \end{array} \right)  \\ \\
     & = & ReLU \left( \begin{array}{c}
         -0.342  \\ 1.92 \\ -1.608 \\
    \end{array} \right) \\ \\
    & = &  \left( \begin{array}{c}
         0  \\ 1.92 \\ 0 \\
    \end{array} \right).
\end{array}
\]
Therefore the score (or output) of the network is,
\[
\begin{array}{lcl}
     \Psi(W,f)(x) & = & W_3 \Big( ReLU \big( W_2(ReLU(W_1 x)) \big) \Big)  \\ \\
     & = & W_3 \left( \begin{array}{c}
         0  \\ 1.92 \\ 0 \\
    \end{array} \right) \\ \\
    & = & \left( \begin{array}{c}
         -0.7(1.92)  \\ 0.1(1.92)
    \end{array} \right) \\ \\
    & = & \left( \begin{array}{c}
         -1.344  \\ 0.192 \\
    \end{array} \right). \\
\end{array}
\]
{\color{black}
Here we have computed the network function as a sequence of linear maps followed by point-wise non-linearities. Let us remark that our definition of network function is equivalent to this one since following Definition~\ref{def:actfun} we have
\[
    \begin{array}{l}
        \act(W,f)_a(x) = -1.2, \\
        \act(W,f)_b(x) = 0.3, \\
        \act(W,f)_c(x) = ReLU\left( 0.2(-1.2) - 0.4(0.3) \right) = ReLU(-0.36) = 0, \\
        \act(W,f)_d(x) = ReLU\left( -1.1(-1.2) + 1.0(0.3) \right) = ReLU(1.62) = 1.62, \\
        \act(W,f)_e(x) = ReLU\left( -0.1(-1.2) - 0.2(0.3) \right) = ReLU(0.06) = 0.06,  \\
        \act(W,f)_f(x) = ReLU\left( -0.6(0) - 0.2(1.62) - 0.3(0.06) \right) = ReLU(-0.342) = 0, \\
        \act(W,f)_g(x) = ReLU\left( 0.3(0) + 1.2(1.62) - 0.4(0.06) \right) = 1.92, \\
        \act(W,f)_h(x) = ReLU\left( -0.1(0) - 1.0(1.62) + 0.2(0.06) \right) = ReLU(-1.608) = 0, \\
        \act(W,f)_i(x) = 0.5(0) - 0.7(1.92) + 0.3(0) = -1.344, \\
        \act(W,f)_j(x) = -1.2(0) + 0.1(1.92) - 0.6(0) = 0.192.  \\
    \end{array}
\]
}
Consider now a change of basis for $(W,f)$.  As mentioned in the text, the change of basis for the input and output neurons are set to 1 (i.e., $\tau_a=\tau_b=\tau_i=\tau_j=1$).  As for the six hidden neurons, let's consider the following six change of basis $\tau= \left( \begin{array}{cc}
    -0.2 & 1.0 \\
    0.3 & -1.0 \\
    -1.1 & 0.1 \\
\end{array} \right)$. These change of basis can be applied to the weights following Eq.~(\ref{eq:group_action}).  Since $\tau$ is 1 for the input neurons, the weights of the first layer get transformed as follows
\[
        \tau W_1 = \left( \begin{array}{cc}
            0.2(-0.2) & -0.4(-0.2)  \\
            -1.1(0.3) & 1.0(0.3) \\
            -0.1(-1.1) & -0.2(-1.1)
        \end{array} \right) = \left( \begin{array}{cc}
        -0.04 & 0.08  \\
        -0.33 & 0.3 \\
        0.11 & 0.22 \\
        \end{array} \right).
\]
The weights of the second layer become
\[
    \tau W_2 = \left( \begin{array}{ccc}
        -0.6 \dfrac{1.0}{-0.2} & -0.2 \dfrac{1.0}{0.3} & -0.3 \dfrac{1.0}{-1.1} \\ \\
        0.3 \dfrac{-1.0}{-0.2} & 1.2 \dfrac{-1.0}{0.3} & -0.4 \dfrac{-1.0}{-1.1} \\ \\
        -0.1 \dfrac{0.1}{-0.2} & -1.0 \dfrac{0.1}{0.3} & 0.2 \dfrac{0.1}{-1.1} \\ \\
    \end{array} \right) = \left( \begin{array}{ccc}
         3 & - \dfrac{0.2}{0.3} &  \dfrac{0.3}{1.1} \\ \\
         \dfrac{0.3}{0.2} & - 4 & - \dfrac{0.4}{1.1} \\ \\
         0.05 & - \dfrac{0.1}{0.3} & - \dfrac{0.02}{1.1} \\ \\
    \end{array} \right),
\]
and those of the third layer
\[
    \tau W_3 = \left( \begin{array}{ccc}
        \dfrac{0.5}{1.0} & \dfrac{-0.7}{-1.0} & \dfrac{0.3}{0.1} \\ \\
        \dfrac{-1.2}{1.0} & \dfrac{0.1}{-1.0} & \dfrac{-0.6}{0.1} \\ \\
    \end{array} \right) = \left( \begin{array}{ccc}
        0.5 & 0.7 & 3  \\
        -1.2 & -0.1 & -6 \\
    \end{array} \right).
\]

As for the activations, one has to apply Eq.~(\ref{eq:activation_action}), i.e., $\tau_v ReLU(\frac{x}{\tau_v})$ in our case.  Note that if $\tau_v>0$ then $\tau_v ReLU(\frac{x}{\tau_v})= \frac{\tau_v}{\tau_v} ReLU(x)=ReLU(x)$ which derives from the positive scale invariance of ReLU.  However, if $\tau_v<0$ then $\tau_v ReLU(\frac{x}{\tau_v})=min(x,0)=g(x)$.  Considering the change of basis matrix $\tau$ given before, it derives that $\tau f = \left( \begin{array}{cc}
    g & ReLU \\
    ReLU & g \\
    g & ReLU \\
\end{array} \right)$. 

Let's apply a forward pass on the neural network $(\tau W, \tau f)$ for the same input $x = \left( \begin{array}{c}
     -1.2  \\ 0.3 \\
\end{array} \right)$. On the first layer we have:
\[
\begin{array}{lcl}
     \left( \begin{array}{c}
        g \\ ReLU \\ g
    \end{array} \right) \tau W_1 x & = & \left( \begin{array}{c}
        g \\ ReLU \\ g
    \end{array} \right) \left( \begin{array}{c}
        -0.04(-1.2) + 0.08(0.3) \\
        -0.33(-1.2) + 0.3(0.3) \\
        0.11(-1.2) + 0.22(0.3) \\
    \end{array} \right)  \\ \\
     & = & \left( \begin{array}{c}
        g \\ ReLU \\ g
    \end{array} \right) \left( \begin{array}{c}
        0.072 \\
        0.486 \\
        -0.066 \\
    \end{array} \right) \\ \\
    & = & \left( \begin{array}{c}
        0 \\
        0.486 \\
        -0.066 \\
    \end{array} \right). \\
\end{array}
\]
Therefore, the activation output of the neurons in the first hidden layer is equal to the activation output on the same neurons on the neural network $(W,f)$ times the corresponding change of basis.  

Propagating the signal to the other layer leads to
\[
    \tau W_2 \left( \begin{array}{c}
         0  \\ 0.486 \\ -0.066 \\
    \end{array} \right) = \left( \begin{array}{c}
         - \dfrac{0.2}{0.3} (0.486) + \dfrac{0.3}{1.1} (-0.066)  \\ \\
         -4 (0.486) - \dfrac{0.4}{1.1}(-0.066) \\ \\ - \dfrac{0.1}{0.3}(0.486) - \dfrac{0.02}{1.1}(-0.066) \\
    \end{array} \right) = \left( \begin{array}{c}
        -0.342  \\  -1.92  \\ -0.1608 \\
    \end{array} \right),
\]
and after applying the activation $\left( \begin{array}{c}
     ReLU  \\ g \\ ReLU \\
\end{array} \right)$ we get the vector $\left( \begin{array}{c}
    0 \\ -1.92 \\ 0 \\
\end{array}\right)$. Finally,
\[
    \Psi \Big( (\tau W, \tau f) \Big)(x) = \tau W_3 \left( \begin{array}{c}
    0 \\ -1.92 \\ 0 \\
\end{array}\right) = \left( \begin{array}{c}
    0.7(-1.92)  \\ -0.1(-1.92) \\
\end{array} \right) = \left( \begin{array}{c}
     -1.344  \\ 0.192 \\
\end{array} \right),
\]
which is the same output than the one for $(W,f)$ computed before.  We can also observe that the activation output of each hidden (and output) neuron on $(\tau W,\tau f)$ is equal to the activation output on that same neuron in $(W,f)$ times the change of basis of that neuron, as noted in the proof of Theorem~\ref{thm:netfunc}.

\section{Example of Theorem~\ref{thm:data}} \label{appendix2}

Here we compute an example to illustrate that $\Psi(W,f)(x)=\Psi(\Wxf,1)(1^d)$. We will work with the notation of Appendix~\ref{appendix1} for the ReLU MLP with explicit weight matrices $W_1, W_2$ and $W_3$ and input vector $x$ given by
\[
\begin{array}{cc}
   W_1 = \left( \begin{array}{cc}
        0.2 & -0.4 \\
        -1.1 & 1.0 \\
        -0.1 & -0.2 \\
        \end{array} \right),  &  W_2 = \left( \begin{array}{ccc}
        -0.6 & -0.2 & -0.3 \\
        0.3 & 1.2 & -0.4 \\
        -0.1 & -1.0 & 0.2 \\
        \end{array} \right), \\ \\
    W_3 = \left( \begin{array}{ccc}
        0.5 & -0.7 & 0.3 \\
        -1.2 & 0.1 & -0.6 \\
        \end{array} \right), &  x = \left( \begin{array}{c}
        -1.2  \\ 0.3 \\
        \end{array} \right).
\end{array}
\]
Recall the definition of the representation $\Wxf$,
\begin{equation*}
    \left( \Wxf \right)_\epsilon = \left\{    \begin{array}{ll}
        W_\epsilon x_{s(\epsilon)} & \text{ if } s(\epsilon) \text{ is an input vertex,}  \\
        W_\epsilon & \text{ if } s(\epsilon) \text{ is a bias vertex,} \\
        W_\epsilon \dfrac{\act(W,f)_{s(\epsilon)}(x) }{  \displaystyle\sum_{\beta \in \zeta_{s(\epsilon)}}  W_\beta \cdot \act(W,f)_{s(\beta)}(x)} & \text{ if } s(\epsilon) \text{ is a hidden vertex. } \\
    \end{array} \right.
\end{equation*}
Denote by $V_1, V_2$ and $V_3$ the weight matrices of $\Wxf$.  We can easily see that $V_1$ is given by
\[
    V_1 = \left( \begin{array}{cc}
        0.2(-1.2) & -0.4(0.3) \\
        -1.1(-1.2) & 1.0(0.3) \\
        -0.1(-1.2) & -0.2(0.3) \\
        \end{array} \right) = \left( \begin{array}{cc}
        -0.24 & -0.12 \\
        1.32 & 0.3 \\
        0.12 & -0.06 \\
        \end{array} \right),
\]
As for the next weight matrices, we have already computed the pre-activations and post-activations of each neuron in a forward pass of $x$ through the network $(W,ReLU)$ in the last appendix, then
\[
    V_2 = \left( \begin{array}{ccc}
        -0.6 \dfrac{0}{-0.36} & -0.2 \dfrac{1.62}{1.62}& -0.3 \dfrac{0.06}{0.06} \\ \\
        0.3 \dfrac{0}{-0.36} & 1.2 \dfrac{1.62}{1.62} & -0.4 \dfrac{0.06}{0.06} \\ \\
        -0.1 \dfrac{0}{-0.36} & -1.0 \dfrac{1.62}{1.62} & 0.2 \dfrac{0.06}{0.06} \\ \\
        \end{array} \right) = \left( \begin{array}{ccc}
        0 & -0.2 & -0.3 \\
        0 & 1.2 & -0.4 \\
        0 & -1.0 & 0.2 \\
        \end{array} \right),
\]
and also
\[
    V_3 = \left( \begin{array}{ccc}
        0.5 \dfrac{0}{-0.342} & -0.7 \dfrac{1.92}{1.92}  & 0.3 \dfrac{0}{-1.608} \\ \\
        -1.2 \dfrac{0}{-0.342} & 0.1 \dfrac{1.92}{1.92} & -0.6 \dfrac{0}{-1.608} \\ \\
        \end{array} \right) = \left( \begin{array}{ccc}
        0 & -0.7  & 0 \\
        0 & 0.1 & 0 \\
        \end{array} \right).
\]
Let's compute a forward pass of the network $(V,1)=(\Wxf,1)$ given the input $\left( \begin{array}{c} 1  \\ 1 \\
\end{array} \right)$ and verify that the output is the same as that of Appendix\ref{appendix1}. We have that
\[
    \begin{array}{lcl}
        V_1 \left( \begin{array}{c} 1  \\ 1 \\
\end{array} \right)  & = & \left( \begin{array}{c}
     -0.24-0.12  \\ 1.32+0.3 \\ 0.12-0.06 \\
\end{array}  \right)  \\ \\
         & = & \left( \begin{array}{c}
             -0.36  \\ 1.62 \\ 0.06 \\
         \end{array} \right), \\
    \end{array}
\]
and
\[
    \begin{array}{lcl}
     V_2 \left( \begin{array}{c}
             -0.36  \\ 1.62 \\ 0.06 \\
         \end{array} \right)  & = &    \left( \begin{array}{c}
             -0.2(1.62)-0.3(0.06)  \\ 1.2(1.62) -0.4(0.06) \\ -1.0(1.62)+0.2(0.06) \\
         \end{array} \right) \\ \\
         & = &  \left( \begin{array}{c}
               -0.342 \\ 1.92 \\ -1.608
         \end{array} \right), \\
    \end{array}
\]
and finally
\[
    \begin{array}{lcl}
        V_3  \left( \begin{array}{c}
                -0.342 \\ 1.92 \\ -1.608
         \end{array} \right) & = & \left( \begin{array}{c}
               -0.7(1.92) \\ 0.1(1.92) \\
         \end{array} \right)  \\
         & = &  \left( \begin{array}{c}
               -1.344 \\ 0.192 \\
         \end{array} \right).
    \end{array}
\]
As noted in the proof of Theorem~\ref{thm:data}, the activation output of each neuron in $(\Wxf,1)$ after a forward pass of the input vector $ \left( \begin{array}{c} 1  \\ 1 \end{array} \right)$, is equal to the pre-activation of that same neuron in $(W,ReLU)$ after a forward pass of $x$.

\section{Double-framed quiver moduli} \label{appendix3}

Let $Q=(\V,\mathcal{E},s,t)$ be a network quiver. Recall that the delooped quiver $Q^\circ=(\V^\circ, \mathcal{E}^\circ,s^\circ,t^\circ)$ is obtained from $Q$ by removing all loops. The hidden quiver $\Q=(\VV,\widetilde{\mathcal{E}},\widetilde{s},\widetilde{t})$ is obtained from the delooped quiver by removing the input and the output layers. Once the complex vector spaces $\Cmplx$ are fixed to every vertex of $Q$, the space of stable thin representations of $Q^\circ$ is 
\[
    \mathcal{R} := \{ W : W_\alpha \in \Cmplx^* \text{ for every } \alpha \in \mathcal{E}^\circ \}.
\]
This is an affine space isomorphic to $\mathcal{R} \cong \Cmplx^n$, where $n$ is the number of elements of $\mathcal{E}^\circ$. The change of basis group we consider here is the group
\[
   \Gt := \displaystyle\prod_{v \in \VV} \Cmplx^*,
\]
whose action on $\mathcal{R}$ is given by Eq.~(\ref{eq:group_action}) in page \pageref{eq:group_action}, i.e., $(\tau\cdot W)_\alpha = \tau_{s(\alpha)}^{-1} W_\alpha \tau_{t(\alpha)}$.
The action (Def.~\ref{def:actiongroup}) of the group $\Gt$ is \textbf{free} if given $g,h \in \Gt$ the existence of an element $x$ with $g \cdot x = h \cdot x$ implies that $g=h$. 

\begin{Lemma}
    The action of the group $\Gt$ is free.
\end{Lemma}

\proof Let $\tau=(\tau_v)_{v \in \VV} \in \Gt$ be an element different from the identity, that is, there is $v \in \VV$ such that $\tau_v\not=1$, and let $W$ be a thin representation. Since $v$ is a vertex of the hidden quiver, there exists an edge $\alpha$ with target $v$ and source $s(\alpha)=u$. If $u$ is a source vertex, then the weight of $\tau \cdot W$ in the edge $\alpha$ is $W_\alpha \tau_v$, which is different from $W_\alpha$ since $W_\alpha\not=0$. If $u$ is not a source, then the weight of $g\cdot W$ in the edge $\alpha$ is $\tau_u^{-1} W_\alpha \tau_v$ which is different from $W_\alpha$ unless $\tau_u=\tau_v$. We can apply the same argument to the vertex $u$ until we reach a source of $Q$, which shows that $\tau \cdot W \not= W$ for any $\tau$ that is not the identity of $\Gt$. This is equivalent to the action of the group being free \citep{Rotman95}.\hfill\BlackBox

Given the action of a group $G$ on a set $X$, the \textbf{$G$-orbit} of an element $x \in X$ is by definition the set $\{ g \cdot x \in X : g \in G \}$. In our case, the $\Gt$-orbit of a thin representation $W$ is the set
\[
    \{ \tau \cdot W : \tau \in \Gt \},
\]  
which is the isomorphism class of the representation $W$, i.e., the set of all representations isomorphic to $W$. From Section 7 we get that the moduli space is the set of all $\Gt$-orbits of elements in $\mathcal{R}$. We will use this to prove the following:

\begin{Theorem}
    The dimension of the moduli space is
    \[
        dim_\Cmplx \left( _d\mathcal{M}_k(\Q) \right) = \# \mathcal{E}^\circ - \# \VV.
    \]
\end{Theorem}

\proof Let $W$ be a thin representation of $Q^\circ$ with non-zero weights. For every hidden vertex $v \in \VV$ we are going to choose once and for all an oriented edge, that we denote by $\alpha_v: v' \to v$. The collection of all the chosen edges $\alpha_v$ and vertices that are targets and sources of them form a subquiver of $Q^\circ$, that we denote $Q^\vee$. The number of hidden vertices of $Q$ and the number of edges in the quiver $Q^\vee$ are the same, by construction. Also, there can not be non-oriented cycles in the quiver $Q^\vee$ since we would have to had chosen two oriented edges with the same target, and we are only choosing one in our construction. This implies that $Q^\vee$ is a union of trees, and the intersection of any two of those trees can only be a source vertex of $Q$ by the same argument. Furthermore, for any of those trees the only vertex that is not a hidden vertex is a unique source of $Q$ corresponding to that tree.

We will show that we can choose a change of basis for each of these trees so that all its weights can be set to $1$ in the isomorphism class of $W$, i.e., the $\Gt$-orbit of that representation. Once this is done, we only have to count how many free choices we have left for weights of $Q^\circ$ that have not been set to $1$. This is exactly the number of oriented edges of $Q^\circ$ minus the number of hidden vertices (which are in correspondence with the oriented edges forming the trees in $Q^\vee$). This will be the dimension of the space of $\Gt$-orbits (i.e., the moduli space) because the previous lemma implies that this is indeed the minimum number of weights to describe the representation $W$ up to isomorphisms.

Let $T$ be a tree in $Q^\vee$ and let $v$ be its source vertex. A change of basis $\tau \in \Gt$ has $\tau_v=1$. Let $\alpha_1:v \to v_1$ be an oriented edge of $T$ with source $v$, and take $\tau_{v_1}=\dfrac{1}{W_{\alpha_1}}$, so after applying $\tau$ to $W$ we obtain that $\left( \tau \cdot W \right)_{\alpha_1} = 1$. Let $\alpha_2:v_1 \to v_2$ be an oriented edge in $T$ that starts in $v_1$, and take $\tau_{v_2}=\dfrac{1}{W_{\alpha_1} W_{\alpha_2}}$. After applying $\tau$ we obtain $\left( \tau \cdot W \right)_{\alpha_2} = 1$. An induction argument shows that up to isomorphism we can take all the weights of $W$ along the tree $T$ to be equal to $1$. Finally, the trees in $Q^\vee$ do not share any oriented edges, therefore they also do not share any vertices except for the source vertices, for which the change of basis is set to $1$. This means that we have chosen a change of basis for every hidden vertex so that the resulting isomorphic representations has all its weights along the trees of $Q^\vee$ equal to 1, which finishes the proof. \hfill\BlackBox

\vskip 0.2in

{\color{black}
\section{Glossary} \label{glossary}

Here we gather all definitions given in this paper alphabetically.

\textbf{Activation function.}
    An \textbf{activation function} is a one variable non-linear function $f: \Cmplx \to \Cmplx$ differentiable except in a set of measure zero.
    
\textbf{Activation output of vertices/neurons.}
    Let $(W,f)$ be a neural network over a network quiver $Q$ and let $x \in \Cmplx^d$ be an input vector of the network. Denote by $\zeta_v$ the set of edges of $Q$ with target $v$. The \textbf{activation output of the vertex} $v \in \V$ \textbf{with respect to} $x$ after applying a forward pass is denoted $\act(W,f)_v(x)$ and is computed as follows:
    \begin{itemize}
        \item If $v\in \V$ is an input vertex, then $\act(W,f)_v(x) = x_v$.
        \item If $v \in \V$ is a bias vertex, then $\act(W,f)_v(x)=1$.
        \item If $v \in \V$ is a hidden vertex, then $\act(W,f)_v(x) = f_v \left( \displaystyle\sum_{\alpha \in \zeta_v} W_\alpha \act(W,f)_{s(\alpha)}(x)  \right)$.
        \item If $v\in \V$ is an output vertex, then $\act(W,f)_v(x) = \displaystyle\sum_{\alpha \in \zeta_v} W_\alpha \act(W,f)_{s(\alpha)} (x)$.
        {\color{black}
        \item If $v\in \V$ is a max-pooling vertex, then $\act(W,f)_v(x) = \maxx_\alpha Re\big( W_{\alpha} \act(W,f)_{s(\alpha)}(x)  \big)$, where $Re$ denotes the real part of a complex number, and the maximum is taken over all $\alpha\in \mathcal{E}$ such that $t(\alpha)=v$.
        }
    \end{itemize}
    
\textbf{Architecture of a neural network.} \citep[page~193]{Goodfellow-et-al-2016}
    The \textbf{architecture} of a neural network refers to its structure which accounts for how many units (neurons) it has and how these units are connected together.

\textbf{Change of basis group of thin representations.}
    The \textbf{change of basis group} of thin representations over a quiver $Q$ is
    \[
    G = \displaystyle\prod_{v \in \V} \Cmplx^*,
    \]
    where $\Cmplx^*$ denotes the multiplicative group of non-zero complex numbers. That is, the elements of $G$ are vectors of non-zero complex numbers $\tau = (\tau_1,...,\tau_n)$ indexed by the set $\V$ of vertices of $Q$, and the group operation between two elements $\tau = (\tau_1,...,\tau_n)$ and $\sigma = (\sigma_1,...,\sigma_n)$ is by definition
    \[
        \tau \sigma := (\tau_1 \sigma_1,..., \tau_n \sigma_n).
    \]  
    
\textbf{Change of basis group of double-framed thin quiver representations.}
    The group of \textbf{change of basis of double-framed thin quiver representations} is the same group $\Gt$ of change of basis of neural networks.
    
\textbf{Change of basis group of neural networks.}
    The \textbf{group of change of basis} for neural networks is denoted as
    \[
        \Gt = \displaystyle\prod_{v \in \VV} \Cmplx^*.
    \]
    An element of the change of basis group $\Gt$ is called a \textbf{change of basis} of the neural network $(W,f)$.
    
\textbf{Co-framed quiver representation.}~\citep{Reineke08}
    A choice of a thin representation $\widetilde{W}$ of the hidden quiver and a map $\ell_v:U_v \to \widetilde{W}_v$ for each $v\in \VV$ determines a pair $(\widetilde{W},\ell)$, where $\ell=\{ \ell_v \}_{v \in \VV}$ that is known as a \textbf{co-framed} quiver representation of $\Q$ by the family of vector spaces $\{ U_v \}_{v \in \VV}$.
    
\textbf{Combinatorial/weight/activation architectures.}
    The \textbf{combinatorial architecture} of a neural network {\color{black}is its} network quiver. The \textbf{weight architecture} is given by constraints on how the weights are chosen, and the \textbf{activation architecture} is the set of activation functions assigned to the loops of the network quiver.
    
\textbf{Data quiver representations.}
    Let $(W,f)$ be a neural network over the network quiver $Q$ and $x\in \Cmplx^d$ an input vector. The thin quiver representation $\Wxf$ is defined as
    \[
    \left( \Wxf \right)_\epsilon = \left\{    \begin{array}{ll}
        W_\epsilon x_{s(\epsilon)} & \text{ if } s(\epsilon) \text{ is an input vertex,}  \\
        W_\epsilon & \text{ if } s(\epsilon) \text{ is a bias vertex,} \\
        W_\epsilon \dfrac{\act(W,f)_{s(\epsilon)}(x) }{  \displaystyle\sum_{\beta \in \zeta_{s(\epsilon)}}  W_\beta \cdot \act(W,f)_{s(\beta)}(x)} & \text{ if } s(\epsilon) \text{ is a hidden vertex, } \\
    \end{array} \right.
    \]
    
\textbf{Delooped quiver of a network quiver.}
    The \textbf{delooped} quiver $Q^\circ$ of $Q$ is the quiver obtained by removing all loops of $Q$. We denote $Q^\circ = (\V, \mathcal{E}^\circ, s^\circ, t^\circ)$.

\textbf{Double-framed thin quiver representation.}
    A \textbf{double-framed} thin quiver representation is a triple $(\ell,\widetilde{W},h)$ where $\widetilde{W}$ is a thin quiver representation of {\color{black}the hidden quiver, $(\widetilde{W},h)$ is a framed representation of $\Q$ and $(\widetilde{W},\ell)$ is a co-framed representation of $\Q$.} 
    
\textbf{Framed quiver representation.}~\citep{Reineke08}
    A choice of a thin representation $\widetilde{W}$ of the hidden quiver and a map $h_v:\widetilde{W}_v \to V_v$ for each $v \in \VV$ determines a pair $(\widetilde{W},h)$, where $h=\{ h_v \}_{v \in \VV}$, that is known as a \textbf{framed} quiver representation of $\Q$ by the family of vector spaces $\{ V_v \}_{v \in \VV}$.

\textbf{Group.} \citep[chap.~1]{Rotman95}
    A non-empty set $G$ is called a \textbf{group} if there exists a function %\newline 
    $\cdot : G \times G \to G$, called the product of the group denoted $a \cdot b$, such that
    \begin{itemize}
        \item $(a \cdot b) \cdot c = a \cdot (b \cdot c)$ for all $a,b,c \in G$.
        \item There exists an element $e \in G$ such that $e\cdot a = a \cdot e = a$ for all $a\in G$, called the \textbf{identity} of $G$.
        \item For each $a\in G$ there exists $a^{-1} \in G$ such that $a \cdot a^{-1} = a^{-1} \cdot a = e$.
    \end{itemize}

\textbf{Group action.} \citep[chap.~3]{Rotman95}
    Let $G$ be a group and let $X$ be a set. We say that there is an \textbf{action of} G \textbf{on} X if there exists a map $\cdot: G \times X \to X$ such that
    \begin{itemize}
        \item $e \cdot x = x$ for all $x\in X$, where $e\in G$ is the identity.
        \item $a \cdot (b \cdot x) = (a b) \cdot x$, for all $a,b \in G$ and all $x\in X$.
    \end{itemize}
    
\textbf{Hidden quiver.}
     The \textbf{hidden quiver} of $Q$, denoted by $\Q = (\VV, \widetilde{\mathcal{E}}, \widetilde{s}, \widetilde{t})$, is given by the hidden vertices $\VV$ of $Q$ and all the oriented edges $\widetilde{\mathcal{E}}$ between hidden vertices of $Q$ that are not loops.
     
\textbf{Input/Output vertices.}
    We call \textbf{input vertices} of the hidden quiver $\Q$ the vertices that are connected to the input vertices of $Q$, and we call \textbf{output vertices} of the hidden quiver $\Q$ the vertices that are connected to the output vertices of $Q$. 

\textbf{Isomorphic quiver representations.}
    Let $Q$ be a quiver and let $W$ and $U$ be two representations of $Q$. If there is a morphism of representations $\tau : W \to U$ where each $\tau_v$ is an invertible linear map, then $W$ and $U$ are said to be \textbf{isomorphic representations}.
    
\textbf{Labeled data set.}
    %Let $Q$ be a network quiver. 
    A \textbf{labeled data set} % for $Q$ 
    is given by a finite set $D = \{ (x_i,t_i) \}_{i=1}^n$ of pairs such that $x_i \in \Cmplx^d$ is a data vector (could also be a matrix or a tensor) and $t_i$ is a target. We can have $t_i \in \Cmplx^k$ for a regression and $t_i \in \{C_0,C_1,...,C_k\}$ for a classification.
    
\textbf{Moduli space.}
    The \textbf{moduli space} of stable double-framed thin quiver representations of $\Q$ is by definition
    \[
        _d\mathcal{M}_k(\Q) := \big\{ [V] : V \in  \ _d\mathcal{R}_k(\Q) \text{ is stable} \big\}.
    \]
    
\textbf{Morphism/Isomorphism of neural networks.}
    Let $(W,f)$ and $(V,g)$ be neural networks over the same network quiver $Q$. A \textbf{morphism of neural networks} $\tau:(W,f) \to (V,g)$ is a morphism of thin quiver representations $\tau:W \to V$ such that $\tau_v=1$ for all $v \in \V$ that is not a hidden vertex, and for every hidden vertex $v \in \V$ the following diagram is commutative
    \[
        \begin{tikzpicture}
         	\matrix (m) [matrix of math nodes,row sep=2em,column sep=2em]
          	{
           	  	\Cmplx  & \Cmplx \\
           	  	\Cmplx & \Cmplx.  \\
           	};
          	\path[-stealth]
           	 (m-1-1) edge node [above] {$f_v$} (m-1-2)
           	 (m-1-2) edge node [right] {$\tau_v$} (m-2-2)
            
           	 (m-1-1) edge node [left] {$\tau_v$} (m-2-1)
           	 (m-2-1) edge node [below] {$g_v$} (m-2-2);
	    \end{tikzpicture}
    \]
    A morphism of neural networks $\tau:(W,f) \to (V,g)$ is an \textbf{isomorphism of neural networks} if $\tau:W \to V$ is an isomorphism of quiver representations.
    We say that two neural networks over $Q$ are \textbf{isomorphic} if there exists an isomorphism of neural networks between them.

\textbf{Morphism of quiver representations.}  \citep[chap.~3]{Assem06}
    Let $Q$ be a quiver and let $W$ and $U$ be two representations of $Q$. A \textbf{morphism of representations} $\tau : W \to U$ is a set of linear maps $\tau = (\tau_v)_{v\in \V}$ indexed by the vertices of $Q$, where $\tau_v:W_v \to U_v$ is a linear map such that $\tau_{t(\epsilon)} W_\epsilon = U_\epsilon \tau_{s(\epsilon)}$ for every $\epsilon \in \E$.
    
\textbf{Network function.}
    Let $(W,f)$ be a neural network over a network quiver $Q$. The \textbf{network function} of the neural network is the function 
    \[
        \Psi(W,f):\Cmplx^d \to \Cmplx^k
    \]
    where the coordinates of $\Psi(W,f)(x)$ are the activation outputs of the output vertices of $(W,f)$ (often called the ``score" of the neural net) with respect to an input vector $x \in \Cmplx^d$.
    
\textbf{Network quiver.}
    A \textbf{network quiver} $Q$ is a quiver arranged by layers such that:
    \begin{itemize}
        \item[1.] There are no loops on source (i.e., input and bias) nor sink vertices.
        \item[2.] There is exactly one loop on each hidden vertex.
        %\item[3.] Other than these loops, there are no more oriented cycles.
    \end{itemize}
    
\textbf{Neural network.}
    A \textbf{neural network} over a network quiver $Q$ is a pair $(W,f)$ where $W$ is a thin representation of the delooped quiver $Q^\circ$ and $f=(f_v)_{v \in \V}$ are activation functions, assigned to the loops of $Q$.

\textbf{Quiver.} \citep[chap.~2]{Assem06}
    A \textbf{quiver} $Q$ is given by a tuple $(\V,\mathcal{E},s,t)$ where  $(\mathcal{V},\mathcal{E})$ is an oriented graph with a set of vertices $\mathcal{V}$ and a set of oriented edges $\mathcal{E}$, and maps $s,t:\mathcal{E} \to \mathcal{V}$ that send $\epsilon \in \mathcal{E}$ to its source vertex $s(\epsilon) \in \mathcal{V}$ and target vertex $t(\epsilon)\in \mathcal{V}$, respectively.
    
\textbf{Quiver arranged by layers.}
    A quiver $Q$ is \textbf{arranged by layers} if it can be drawn from left to right arranging its vertices in columns such that:
    \begin{itemize}
        \item There are no oriented edges from vertices on the right to vertices on the left.
        \item There are no oriented edges between vertices in the same column, other than loops and edges from bias vertices.
    \end{itemize}
    The first layer on the left, called the \textbf{input layer}, will be formed by the $d$ input vertices. The last layer on the right, called the \textbf{output layer}, will be formed by the $k$ output vertices. The layers that are not input nor output layers are called \textbf{hidden layers}.  We enumerate the hidden layers from left to right as : $1^{\mbox{\scriptsize st}}$ hidden layer, $2^{\mbox{\scriptsize nd}}$ hidden layer, $3^{\mbox{\scriptsize rd}}$ hidden layer, and so on.

\textbf{Quiver representation.}  \citep[chap.~3]{Assem06}
    If $Q$ is a quiver, a \textbf{quiver representation} of $Q$ is given by a pair of sets 
    \[
        W := \big( (W_v)_{v\in \V}, (W_\epsilon)_{\epsilon \in \E} \big)
    \]
    where the $W_v$'s are vector spaces indexed by the vertices of $Q$, and the $W_\epsilon$'s are linear maps indexed by the oriented edges of $Q$, such that for every edge $\epsilon \in \E$
    \[
        W_\epsilon : W_{s(\epsilon)} \to W_{t(\epsilon)}.
    \]
    
\textbf{Representation space.}
    The \textbf{representation space} $_d\mathcal{R}_k(\Q)$ of the hidden quiver $\Q$ of a network quiver $Q$, is the set of all possible double-framed thin quiver representations of $\Q$.
    
\textbf{Source/Sink vertices.} \cite[chap.~2]{Assem06}
      A \textbf{source vertex} of a quiver $Q$ is a vertex $v\in \V$ such that there are no oriented edges $\epsilon \in \mathcal{E}$ with target $t(\epsilon)=v$. A \textbf{sink vertex} of a quiver $Q$ is a vertex $v \in \V$ such that there are no oriented edges $\epsilon \in \mathcal{E}$ with source $s(\epsilon)=v$. A \textbf{loop} in a quiver $Q$ is an oriented edge $\epsilon$ such that $s(\epsilon)=t(\epsilon)$.
      
\textbf{Stable quiver representation.}
    A double-framed thin quiver representation $(\ell,\widetilde{W},h)$ is \textbf{stable} if the following two conditions are satisfied:
    \begin{itemize}
        \item[1.] The only sub-representation $U$ of $\widetilde{W}$ which is contained in $ker(h)$ is the zero sub-representation, and
        \item[2.] The only sub-representation $U$ of $\widetilde{W}$ that contains $Im(\ell)$ is $\widetilde{W}$.
    \end{itemize}

\textbf{Sub-representation.} \citep[page~14]{Schiffler14} 
    Let $W$ be a thin representation of the delooped quiver $Q^\circ$ of a network quiver $Q$. A \textbf{sub-representation} of $W$ is a representation $U$ of $Q^\circ$ such that there is a morphism of representations $\tau:U \to W$ where each map $\tau_v$ is an injective map.

\textbf{Teleportation.}
    Let $(W,f)$ be a neural network and let $\tau \in \Gt$ be an element of the group of change of basis of neural networks such that the isomorphic neural network $\tau \cdot (W,f)$ has the same weight architecture as $(W,f)$. The \textbf{teleportation} of the neural network $(W,f)$ with respect to $\tau$ is the neural network $\tau \cdot (W,f)$.  

\textbf{Thin quiver representation.}
    A \textbf{thin representation} of a quiver $Q$ is a quiver representation $W$ such that $W_v=\Cmplx$ for all $v \in V$.

\textbf{Zero representation.}
    The \textbf{zero representation} of $Q$ is the representation denoted $0$ where every vector space assigned to every vertex is the zero vector space, and therefore every linear map in it is also zero.
}

\bibliography{ref}

\end{document}